\documentclass[utf8]{frontiersSCNS} 

\usepackage{url,microtype,subcaption}
\usepackage[onehalfspacing]{setspace}

\usepackage{amssymb}
\usepackage{latexsym}

\usepackage[draft]{hyperref} 

\usepackage{amsmath}
\usepackage{placeins}
\usepackage[american]{babel}
\usepackage{textcomp}
\usepackage{gensymb}
\usepackage{dirtytalk}

\newcommand{\mat}[1]{\uppercase{\boldsymbol{#1}}}
\renewcommand{\vec}[1]{\lowercase{\boldsymbol{#1}}}
\DeclareMathOperator*{\argmax}{\arg\!\max}

\newcommand{\rebmod}[1]{#1
}
\newcommand{\rebnew}[1]{#1
}

\usepackage{lipsum}

\newcommand\blfootnote[1]{\begingroup
  \renewcommand\thefootnote{}\footnote{#1}\addtocounter{footnote}{-1}\endgroup
}

\def\keyFont{\fontsize{8}{11}\helveticabold }
\def\firstAuthorLast{Gruijthuijsen, Garcia-Peraza-Herrera {et~al.}} \def\Authors{Caspar Gruijthuijsen\,$^{1\dagger}$, 
Luis C. Garcia-Peraza-Herrera\,$^{2, 4*\dagger}$,
Gianni Borghesan\,$^{1}$,
Dominiek Reynaerts\,$^{1}$,
Jan Deprest\,$^{3}$,
Sebastien Ourselin\,$^{4}$,
Tom Vercauteren\,$^{4}$,
and Emmanuel Vander Poorten\,$^{1}$}
\def\Address{$^{1}$Department of Mechanical Engineering, KU Leuven, Leuven, Belgium \\
$^{2}$Department of Medical Physics and Biomedical Engineering, University College London, London, United Kingdom \\
$^{3}$Department of Development and Regeneration, Division Woman and Child, KU Leuven, Leuven, Belgium \\
$^{4}$Department of Surgical \& Interventional Engineering, King's College London, London, United Kingdom}
\def\corrAuthor{9th Floor, Becket House, 1 Lambeth Palace Road, London, SE1 7EU
}

\def\corrEmail{\href{mailto:luis\_c.garcia\_peraza\_herrera@kcl.ac.uk}{luis\_c.garcia\_peraza\_herrera@kcl.ac.uk}}

\begin{document}
\onecolumn
\firstpage{1}

\title[Autonomous Instrument Tracking]{Robotic Endoscope Control via Autonomous Instrument Tracking} 

\author[\firstAuthorLast ]{\Authors} \address{} \correspondance{} 

\extraAuth{}

\maketitle

\begin{abstract}Many keyhole interventions rely on bi-manual handling of surgical instruments,
forcing the main surgeon to rely on a second surgeon to act as a camera assistant.
In addition to the burden of excessively involving surgical staff, this may lead to reduced image stability, increased task completion time and sometimes errors
due to the monotony of the task.
Robotic endoscope holders, controlled by a set of basic instructions, have been proposed as an alternative, but their unnatural handling may increase the cognitive load of the (solo) surgeon, which hinders their clinical acceptance. 
More seamless integration in the surgical workflow would be achieved if robotic endoscope holders collaborated with the operating surgeon via semantically rich instructions that closely resemble instructions that would otherwise be issued to a human camera assistant, such as "focus on my right-hand instrument". As a proof of concept, this paper presents a novel system that paves the way towards a synergistic interaction between surgeons and robotic endoscope holders.
The proposed platform allows the surgeon to perform a bimanual coordination and navigation task, while a robotic arm autonomously performs the endoscope positioning tasks. Within our system, we propose a novel tooltip localization method based on surgical tool segmentation and a novel visual servoing approach that ensures smooth and appropriate motion of the endoscope camera.
We validate our vision pipeline and run a user study of this system.
The clinical relevance of the study is ensured through the use of a 
laparoscopic exercise validated by the European Academy of Gynaecological Surgery which involves bi-manual coordination and navigation.
Successful application of our proposed system provides a promising starting point towards broader clinical adoption of robotic endoscope holders.
\tiny
 \keyFont{ \section{Keywords:} Minimally Invasive Surgery, Endoscope Holders, Endoscope Robots, Endoscope Control, Visual Servoing} 
\end{abstract} \blfootnote{$^\dagger$ These authors have contributed equally to this work and share first authorship.}
\section{Introduction}
\label{sec:introduction}

In recent years, many surgical procedures shifted from open surgery to minimally invasive surgery (MIS). Although MIS offers excellent advantages for the patient, including reduced scarring and faster recovery, it comes with challenges for the surgical team. Most notable is the loss of direct view onto the surgical site. In keyhole surgery, the surgeon manipulates long and slender instruments introduced into the patient through small incisions or keyholes. The surgeon relies on endoscopes, also long and slender instruments equipped with a camera and light source, to obtain visual feedback on the scene and the relative pose of the other instruments. The limited field of view (FoV) and 
depth of field
of the endoscope urge an efficient endoscope manipulation method that allows covering all the important features and hereto optimizes the view at all times.

In typical MIS, surgeons cannot manipulate the endoscope themselves as their hands are occupied with other instruments. Therefore, a camera assistant, typically
another surgeon
takes charge of handling the endoscope.
Human camera assistants have a number of shortcomings.
An important drawback relates to the cost of the human camera assistant \citep{Stott2017}. Arguably, highly trained clinicians could better be assigned to other surgical duties that require the full extent of their skill set (as opposed to mainly manipulating the endoscope).
If made widely feasible, solo MIS surgery would 
improve cost-effectiveness and staffing efficiency.
An additional source of weakness related to human camera assistants is the ergonomic burden associated with assisting in MIS \citep{Wauben2006,Lee2009}.
This may lead to reduced image stability, fatigue, distractions, increased task completion times, and erroneous involuntary movements \citep{Goodell2006,Platte2019,RodriguesArmijo2020}. This problem aggravates for long interventions or when the assistant has to adopt particularly uncomfortable postures.
Besides the ergonomic challenges, miscommunication between the surgeon and the assistant may lead to sub-optimal views \citep{Amin2020}.

In order to help or bypass the human camera assistant and to optimize image stability, numerous endoscope holders have been designed in the past~\citep{Jaspers2004,Bihlmaier2016,Takahashi:Handbook:2020}. 
One can distinguish passive endoscope holders and active or robotic endoscope holders. 
Passive endoscope holders are mechanical devices that lock the endoscope in a given position until manually unlocked and adjusted. A problem common to passive endoscope holders is that they result in an intermittent operation that interferes with the manipulation task~\citep{Jaspers2004}. When surgeons want to adjust the endoscopic view themselves, they will have to free one or both hands to reposition the endoscope. 
To counter this problem, robotic endoscope holders have been developed. These motorized devices offer the surgeon a dedicated interface to control the endoscope pose. Well-designed robotic endoscope holders do not cause additional fatigue, improve image stability, and increase ergonomics \citep{Fujii2018}. Also, hand-eye coordination issues may be avoided. Overall such robotic endoscope holders may lower the cognitive load of the surgeon and reduce operating room (OR) staff time and intervention cost \citep{Ali2018}. 
However, despite these advantages and the number of systems available, robotic endoscope holders have not found widespread clinical acceptance~\citep{Bihlmaier2016}. This has been linked to the suboptimal nature of the human interface and consequently the discomfort caused to the surgeon by the increased cognitive load needed to control the camera.
Popular robotic endoscope holders use foot pedals, joysticks, voice control, gaze control, and head movements \citep{Kommu2007,Hollander2014,Fujii2018}. The context switching between surgical manipulation and these camera control mechanisms seems to hinder the ability of the surgeon to concentrate on the main surgical task~\citep{Bihlmaier2016}.

\subsection{Contributions}
In this work, we introduce the framework of \emph{semantically rich endoscope control}, which is our proposal on how robotic endoscope control could be implemented to mitigate interruptions and maximize the clinical acceptance of robotic endoscope holders. 
We claim that \emph{semantically rich instructions} that relate to the instruments such as ``focus on the right/left instrument" and ``focus on a point between the instruments" are a priority, as they are shared among a large number of surgical procedures. 
\rebmod{Therefore, we present a novel system that paves the way towards a synergistic interaction between surgeons and robotic endoscope holders. 
To the best of our knowledge, we are the first to report how to construct 
an autonomous instrument tracking system that allows for solo-surgery using only the endoscope as a sensor to track the surgical tools. The proposed platform allows the surgeon to perform a bi-manual coordination and navigation tasks while the robotic arm autonomously performs the endoscope positioning.
}

\rebnew{Within our proposed platform, we introduce a novel tooltip localization method based on a hybrid mixture of deep learning and classical computer vision. In contrast to other tool localization methods in the literature, the proposed approach does not require manual annotations of the tooltips, but relies on tool segmentation, which is advantageous as the manual annotation effort could be trivially waived employing methods such as that recently proposed in \cite{Garcia-Peraza-Herrera2021}.
This vision pipeline was individually validated and the proposed tooltip localization method was able to detect tips in $84.46$\% of the frames. This performance proved sufficient to allow for a successful autonomous guidance of the endoscope (per user study of the whole robotic system).
}

\rebnew{We propose a novel visual servoing method for a generalized endoscope model with support for both remote center of motion and endoscope bending. We show that a hybrid of position-based visual servoing (PBVS) and 3D image-based visual-servoing (IBVS) is preferred for robotic endoscope control. 
}

\rebmod{We run a user study of the whole robotic system on a standardized bi-manual coordination and navigation laparoscopic task accredited for surgical training~\citep{LASTT2020}.
In this study we show that the combination of novel tool localization and visual servoing proposed is
robust enough to allow for the successful autonomous control of the endoscope. 
During the user study experiments ($8$ people, $5$ trials), participants were able to complete the bi-manual coordination surgical task without the aid of a camera assistant and  in a reasonable time ($172\,$s on average). 
}

\subsection{Towards semantically rich robotic endoscope control}

While solo surgery has been demonstrated with simple robotic endoscope control approaches \citep{Takahashi:Handbook:2020}, we argue that to overcome the usability issues that impede broad clinical adoption of robotic endoscope holders and move towards solo surgery, 
robotic endoscope control should be performed at the task autonomy level. 
To efficiently operate in this setting, a robotic endoscope holder should accept a set of \emph{semantically rich instructions}.
These instructions correspond to the commands that a surgeon would normally issue to a human camera assistant. This contrasts with earlier approaches, where the very limited instruction sets (up, down, left, right, zoom in, zoom out) lead to a semantic gap between the robotic endoscopic holder and the surgeon~\citep{Kunze2011}. With semantically rich instructions, it would be possible to bridge this gap and restore the familiar relationship between the surgeon and the (now tireless and precise) camera assistant.

A semantically rich instruction set should contain commands that induce context-aware actions. Examples of such are ``zoom in on the last suture", ``hold the camera stationary above the liver", and ``focus the camera on my right instrument". When these instructions are autonomously executed by a robotic endoscope holder, we refer to the control as \emph{semantically rich robotic endoscope control}.
We believe that semantically rich robotic endoscope control can effectively overcome the problem of intermittent operation with endoscopic holders, does not disrupt the established surgical workflow, ensures minimal overhead for the surgeon, and overall maximizes the usability and efficiency of the intervention.

\begin{figure}[hb!]
    \centering
    \includegraphics[width=0.7\columnwidth]{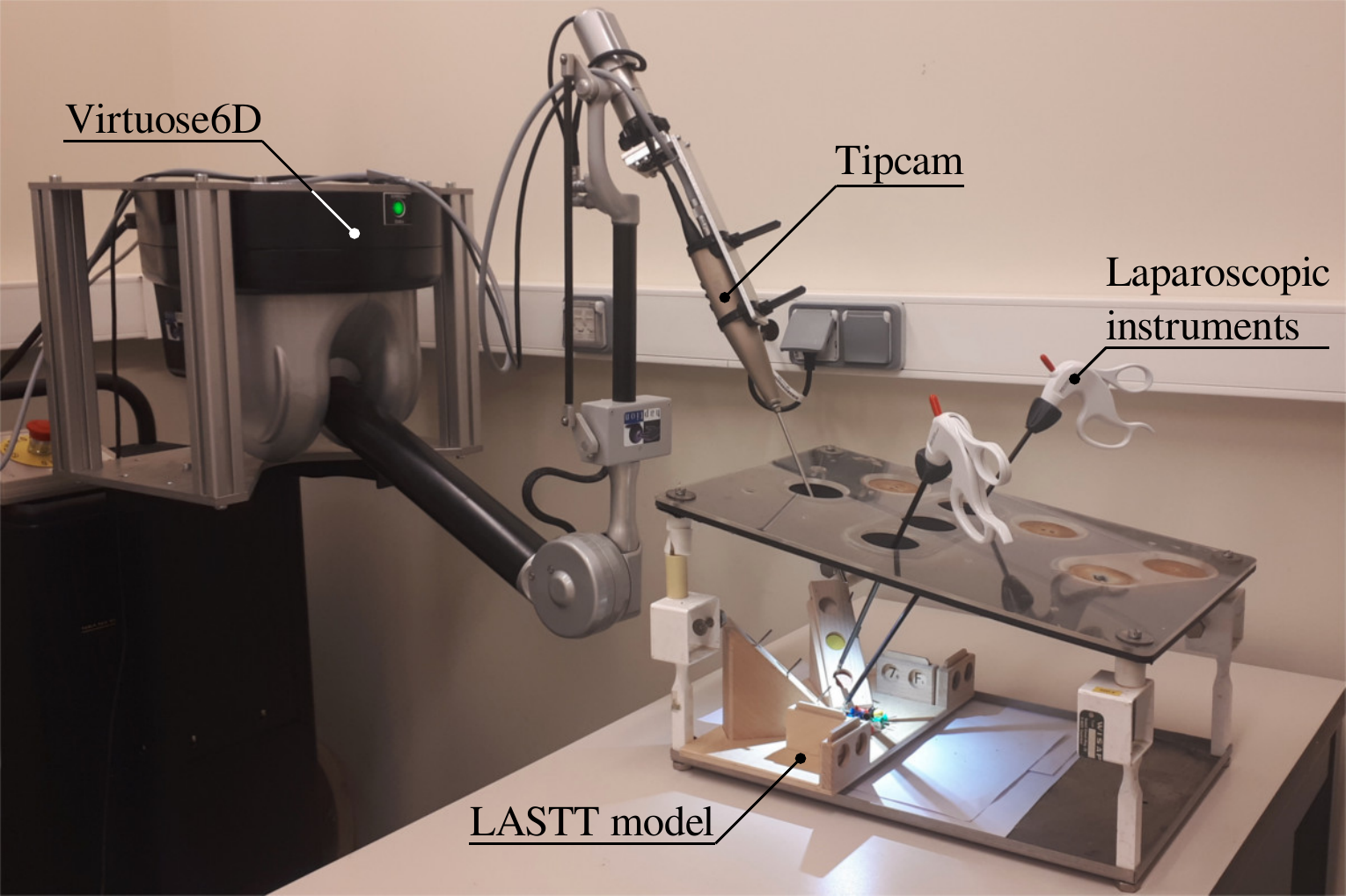}
    \caption{Proposed experimental setup for autonomous endoscope control in a laparoscopic setting. The LASTT model \citep{LASTT2020} showcased within the box trainer is designed for the simulation of simple surgical tasks with a focus on gynaecological procedures. It is common ground for the practice and evaluation of hand-eye and bi-manual coordination skills.}
    \label{fig:endoscope_guidance_setup}
\end{figure}

Although instructions that have the camera track an anatomical feature are relevant, autonomous instrument tracking instructions (e.g. ``focus the camera between the instruments") play a
prominent role, as they are common to a large number of laparoscopic procedures and form a fundamental step towards solo surgery. Therefore, in this work we focus on semantically rich instructions related to the autonomous instrument tracking (AIT) of a maximum of two endoscopic instruments (one per hand of the operating surgeon, see Fig. \ref{fig:endoscope_guidance_setup}). Particularly, the proposed method implements the instructions ``focus on the right/left instrument" and ``focus on a point between the instruments". User interface methods to translate requests expressed by the surgeon (e.g. voice control) to these AIT instructions fall outside the scope of this work.

The remainder of the paper is organized as follows.
After describing the related work, the AIT problem is stated in Sec.~\ref{sec:autonomous_instrument_tracking}.
The quality of the AIT depends on robust methods to localize one or more surgical instruments in the endoscopic view.
Sec.~\ref{sec:instrument_localization} describes a novel image-processing pipeline that was developed to tackle this problem. Visual servoing methods are described in Sec.~\ref{sec:visual_servoing}. These methods provide the robotic endoscope control with the ability to track the detected instruments autonomously. An experimental user study campaign is set up and described in Sec.~\ref{sec:experiments} to demonstrate the value of AIT in a validated surgical training task. Sec.~\ref{sec:results_and_discussion} discusses the obtained results and Sec.~\ref{sec:conclusion} draws conclusions regarding the implementation of the AIT instructions proposed in this work.

 \section{Related work}
\label{sec:related_work}
Robotic endoscope control (REC) allows the surgeon to control the endoscope without having to free their hands. A wide variety of dedicated control interfaces have been developed and commercialized for this purpose, including joystick control, voice control, gaze control and head gesture control~\citep{Taniguchi2010}. 
Despite the apparent differences in these interfaces,
established approaches
offer the surgeon a basic instruction set to control the endoscope. This set typically consists of six instructions: zoom in/out, move up/down, and move left/right. 
The basic nature of these instructions makes detailed positioning cumbersome, usually resorting to a lengthy list of instructions. This shifts the surgeon's focus from handling the surgical instruments towards the positioning of the endoscope, as concurrent execution of those actions is often strenuous and confusing~\citep{Jaspers2004}. 
Moreover, the increased mental workload stemming from simultaneous control of the instruments and the endoscope might adversely affect intervention outcomes~\citep{Bihlmaier2016}. Similar to passive endoscope holders, robotic endoscope holders that offer a basic set of instructions prevent fluid operation and lead to intermittent action.

A number of REC approaches that pursue fully autonomous operation have been proposed as well. A 
starting point for an
autonomous REC strategy is to reposition the endoscope so as to keep the surgical instrument tip centered in the view. Such an approach could work already when only the 2D position of the instrument in the endoscopic image is available~\citep{Uecker1995,Osa2010,Agustinos2014,Zinchenko2021}. In this case, the endoscope zoom level (depth) is left uncontrolled. Some of these methods also require a 3D geometrical instrument model~\citep{Agustinos2014}, limiting the flexibility of the system.
Approaches such as proposed by~\cite{Zinchenko2021} have also suggested to replace the endoscope screen with a head-mounted virtual reality device that facilitates the estimation of the surgeon's attention focus from the headset's gyroscope. In this scenario, the autonomous REC strategy aims to reposition the endoscope with the aim of maintaining the weighted center of mass between the instruments' contour centers and the point of focus in the center of the view.
However, it has been shown in works such as~\citep{Hanna1997,Nishikawa2008} that the zoom level is important for effective endoscope positioning. Other authors tried to circumvent the lack of depth information in 2D endoscopic images by relating the inter-instrument distance to the zoom level~\citep{Song2012,King2013}. This approach is obviously limited to situations where at least two instruments are visible. 

When the 3D instrument tip position is available, smarter autonomous REC strategies are possible.
In the context of fully robotic surgery,
kinematic-based tooltip position information has been used to provide autonomously guided ultrasound imaging with corresponding augmented reality display for the surgeon~\citep{Samei:MedIA:2020}. 
Kinematics have also been employed by Mariani and Da Col~\emph{et~al.}~\citep{Mariani2020,DaCol2021} for autonomous endoscope guidance in a user study on \emph{ex vivo} bladder reconstruction with the da Vinci Surgical System. In their experimental setup, the system could track either a single instrument or the midpoint between two tools.
\rebmod{Similarly, \cite{Avellino2020} have also employed kinematics for autonomous endoscope guidance in a co-manipulation scenario
\footnote{\url{https://www.youtube.com/watch?v=R1qwKAWFOIk}
}.
}
In~\citep{Casals1996,Mudunuri2010}, rule-based strategies switch the control mode between single-instrument tracking or tracking points that aggregate locations of all visible instruments. Pandya~\emph{et~al.} argued that such schemes are reactive and that better results can be obtained with predictive schemes, which incorporate knowledge of the surgery and the surgical phase~\citep{Pandya2014}. Examples of such knowledge-based methods are~\citep{Wang1998,Kwon2008,Weede2011,RivasBlanco2014,Bihlmaier2016,wagner21}. While promising in theory, in practice, the effort to create complete and reliable models for an entire surgery is excessive for current surgical data science systems.
In addition,  accurate and highly robust surgical phase recognition algorithms are required, increasing the complexity of this solution considerably.

\rebmod{With regards to the levels of autonomy in robotic surgery, ~\cite{Yang2017} have recently highlighted that the above strategies aim for very high autonomy levels but take no advantage of the surgeon's presence.
In essence, the surgeon is left with an empty instruction set
to direct the endoscope holder.
Besides being hard to implement given the current state of the art, such high autonomy levels may be impractical and hard to transfer to clinical practice.
Effectively, an ideal camera assistant only functions at the task autonomy level. This is also in line with the recent study by \cite{Col2020}, who concluded that it is important for endoscope control tasks to find the right trade-off between user control and autonomy.
}

\rebnew{To 
facilitate the autonomous endoscope guidance for laparoscopic applications when the 3D instrument tip position is not available, some authors have proposed to attach different types of markers to the instruments (e.g. optical, electromagnetic). This modification often comes with extra sensing equipment that needs to be added to the operating room.
}

\rebnew{In \cite{Song2012a}, authors proposed to use a monocular webcam mounted on a robotic pan-tilt platform to track two laparoscopic instruments with two colored rings attached to each instrument.
They employed the estimated 2D image coordinates of the fiducial markers to control all the degrees of freedom of the robotic platform. However, this image-based visual servoing is not able to attain a desired constant depth to the target tissue (as also shown in our simulation of image-based visual servoing in Sec. \ref{sec:vs_discussion}).
In addition, the choice of fiducial markers is also an issue.
Over the years, going back at least as far as to \citep{Uenohara1995}, many types of markers have been proposed by the community for tool tracking purposes. For example, straight lines \citep{Casals1996}, black stripes \citep{Zhang2002}, cyan rings \citep{Tonet2007}, green stripes \citep{Reiter2011a}, multiple colour rings for multiple instruments (blue-orange, blue-yellow) \citep{Ko2005}, and multiple colour (red, yellow, cyan and green) bio-compatible markers \citep{Bouarfa2012}.
However, although fiducial markers such as colored rings ease the tracking of surgical instruments, attaching or coating surgical instruments with fiducial markers presents serious sterilization, legal and installation challenges \citep{Stoyanov2012,Bouget2017}. 
First, the vision system requires specific tools to work or a modification of the current ones, 
which 
introduces a challenge for clinical translation. 
At the same time, computational methods designed to work with fiducials cannot easily be trained with standard previously recorded interventions. Additionally, to be used in human experiments, the markers need to be robust to the sterilisation process (e.g. autoclave). This poses a manufacturing challenge and increases the cost of the instruments. The positioning of the markers is also challenging. If they are too close to the tip, they might be occluded by the tissue being manipulated. If they are placed back in the shaft, they might be hidden to the camera, as surgeons tend to place the endoscope close to the operating point. Even if they are optimally positioned, fiducials may be easily covered by blood, smoke, or pieces of tissue. In addition to occlusions, illumination (reflections, shadows) and viewpoint changes still remain a challenge for the detection of the fiducial markers.
}

\rebnew{In contrast to using colored markers, \cite{Sandoval2021} used a motion capture system (MoCap) in the operating room to help the autonomous instrument tracking. The MoCap consisted of an exteroceptive sensor composed of $8$ high resolution infrared cameras. This system was able to provide 
the position of the reflective markers placed at the instruments ($4$ markers per instrument) in real time. 
However, the MoCap increases considerably the cost of the proposed system and complicates the surgical workflow. 
Instruments need to be modified to add the markers, and the MoCap needs to be installed in the operating room. 
As any other optical tracking system, it also possesses the risk of occlusions in the line of sight, making it impossible for the system to track the instruments when such occlusions occur. 
As opposed to all these different markers, the endoscope is necessary to perform the surgery, and the surgeon needs to be able to see the instruments to carry out the intervention, so using the endoscope and its video frames without any instrument modifications to help track the tools is a solution that stems naturally from the existing surgical workflow.
This has also been the path followed in the devise of \textsc{AutoLap}™ (Medical Surgery Technologies, Yokneam, Israel) \citep{Wijsman2018,Wijsman2022}, which is, to the best of our knowledge, the only robotic laparoscopic camera holder that claims to have incorporated image-based laparoscopic camera steering within its features. However, no technical details are provided in these publications on how it is achieved. 
}

\section{Autonomous instrument tracking}
\label{sec:autonomous_instrument_tracking}

In a typical surgical scenario, a surgeon manipulates two instruments: one in the dominant and one in the non-dominant hand. In such a case, the surgeon might want to focus the camera on one specific instrument, or center the view on a point in between the instruments, depending on their relative importance. AIT strives to automate these tasks, as explained next.

\subsection{Centering instrument tips in FoV}
\label{sec:centering}

\begin{figure}[tb]
    \centering
    \includegraphics[width=0.35\textwidth]{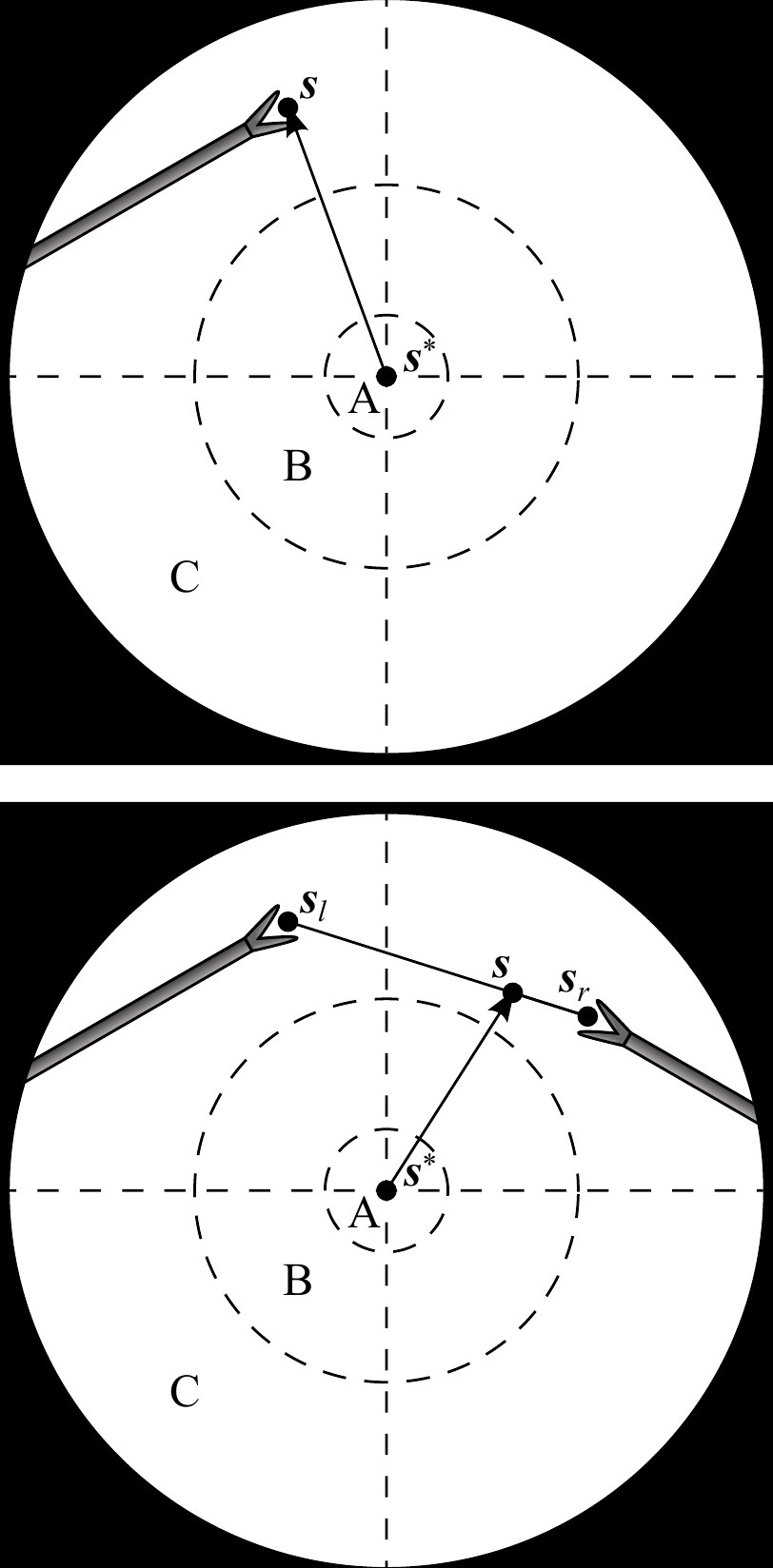}
    \caption{Endoscopic view with one instrument tip at position $\vec{s}$ (top), and two instrument tips at positions $\vec{s}_l$ and $\vec{s}_r$, combined to a virtual tip position $\vec{s}$ (bottom). The AIT controller aims to make $\vec{s}$ coincide with a desired position $\vec{s}^*$. A, B and C are respectively the target, transition and violation zones of a programmed position hysteresis approach.}
    \label{fig:ait_objective}
\end{figure}

With one instrument present, the proposed AIT objective is to center the instrument tip position $\vec{s}$ in the FoV, as is illustrated in Fig.~\ref{fig:ait_objective} (top). With two visible instruments, a relative dominance factor $w_d \in [0, 1]$ can be assigned to the instruments (adjustable via a semantically rich instruction ``change dominance factor X\% to the right/left''). The AIT controller can then track the virtual average tip position according to
\begin{equation}
\label{eq:virtual_instrument_tip}
    \vec{s} = (1-w_d) \vec{s}_l + w_d \vec{s}_r,
\end{equation}
where $\vec{s}_l$ and $\vec{s}_r$ are the respective tip positions of the left and right instrument as visualized in Fig.~\ref{fig:ait_objective}, bottom.

If the AIT were implemented to continuously track the virtual tip position $\vec{s}$, the view would never come to a standstill, which would be disturbing for the surgeon. As a solution, also suggested  in~\citep{Bihlmaier2016} and \cite{eslamian2020development}, a position hysteresis behaviour can be implemented.
In this work, a combination of
instructions with position hysteresis is implemented
based on three zones in the endoscope FoV. As illustrated in
Fig.~\ref{fig:ait_objective},
target zone A captures the ideal location of the tooltip, and transition zone B represents a tolerated region. Entering a violation zone C triggers re-positioning of the endoscope.
Whenever $\vec{s}$ moves from zone B to zone C, the AIT will be activated. It will then stay active until $\vec{s}$ reaches zone A. Afterwards, the FoV will be kept stable, until $\vec{s}$ again crosses the border between zone B and zone C.

This implementation of AIT offers the surgeon instructions to track either instrument, to change the dominance factor, or to stop the tracking by disabling the AIT. Note that this implementation of AIT only specifies two degrees of freedom (DoFs) out of the four available DoFs in typical laparoscopy. 
\rebmod{The depth DoF is controlled by an additional instruction for the zoom level, i.e., the distance between the camera and the instrument tip. The DoF that rolls the endoscope around its viewing axis is controlled to always enforce an intuitive horizontal orientation of the camera horizon. If desired, a semantically rich instruction could be added to alter this behaviour.
}

\subsection{Comanipulation fallback}
\label{sec:comanipulation fallback}
As neither the set of AIT instructions nor any other set of instructions can realistically cover all instructions relevant for semantically rich REC, situations can arise in surgical practice where the capabilities of the robotic endoscope holder are insufficient. In such a case, it is necessary to switch to a comanipulation mode. 
This kind of switching is already the clinical reality for commercial robotic endoscope holders~\citep{Gillen2014,Hollander2014}
and is particularly relevant when the system is used to support rather than replace the surgical assistant.

This work proposes to embed an easy switching functionality as a system feature. A natural transition from REC to comanipulation mode can be made possible through the use of a mechanically backdrivable robotic endoscope holder. 
This way, no extra hardware components are needed for switching, neither is it necessary to release the endoscope from the robotic endoscope holder. Instead, 
the surgeon can simply release one instrument, grab the endoscope and comanipulate it jointly with the robotic endoscope holder. During comanipulation, the human 
provides the intelligence behind the endoscope motions, while still experiencing support in the form of tremor-eliminating damping and fatigue-reducing gravity compensation. 
Such an approach broadens the scope of interventions where REC can be realistically applied. \section{Markerless instrument localization}
\label{sec:instrument_localization}
REC based on semantically rich instructions requires the robotic endoscope holder to autonomously execute context-aware tasks.
This implies a need to autonomously collect contextual information. The AIT instruction relies on knowledge of the tip position $\vec{s}$ of the surgical instruments in the endoscopic view.
To obtain this information, without the need to alter the employed instruments or surgical workflow, a markerless instrument localization pipeline is developed in this section.
Note that the term \emph{localization} is employed here, instead of the commonly used term \emph{tracking}, as for the sake of clarity this work reserves \emph{tracking} for the robotic servoing approaches needed for AIT.

\subsection{Instrument localization approaches}
\label{sec:instrument_localization_approaches}
If, in addition to the endoscope, the instruments are also mounted on a robotic system~\citep{Eslamian2016,Weede2011} or if they are monitored by an external measurement system~\citep{Nishikawa2008,Polski2009}, the position of the instruments can be directly obtained, provided that all involved systems are correctly registered and calibrated. However, in this work, manual handling of off-the-shelf laparoscopic instruments precludes access to such external localization information.

An alternative, which we use in this work, is to exploit the endoscope itself as the sensor. 
\rebnew{A review on this topic has been published relatively recently by \cite{Bouget2017}. In their work Bouget et al. present a comprehensive survey of the last years of research in tool detection and tracking with a particular focus on methods proposed prior to the advent of the deep learning approaches.}
Recent 
instrument localization techniques based on Convolutional Neural Networks (CNN) \citep{Gonzalez2020, Pakhomov2020} are currently recognized as the state-of-the-art
approaches~\citep{Allan2019, Ross:MedIA:2021} for such problems. In this work, we leverage our previous experience with CNN-based real-time tool segmentation networks \citep{Garcia-Peraza-Herrera2017c, Garcia-Peraza-Herrera2017} and embed the segmentation in a stereo pipeline to estimate the location of the tooltips in 3D.

\subsection{Instrument localization pipeline}
\label{sec:instrument_localization_pipeline}
\begin{figure}[hb!]
    \centering
    \includegraphics[width=1.0\columnwidth]{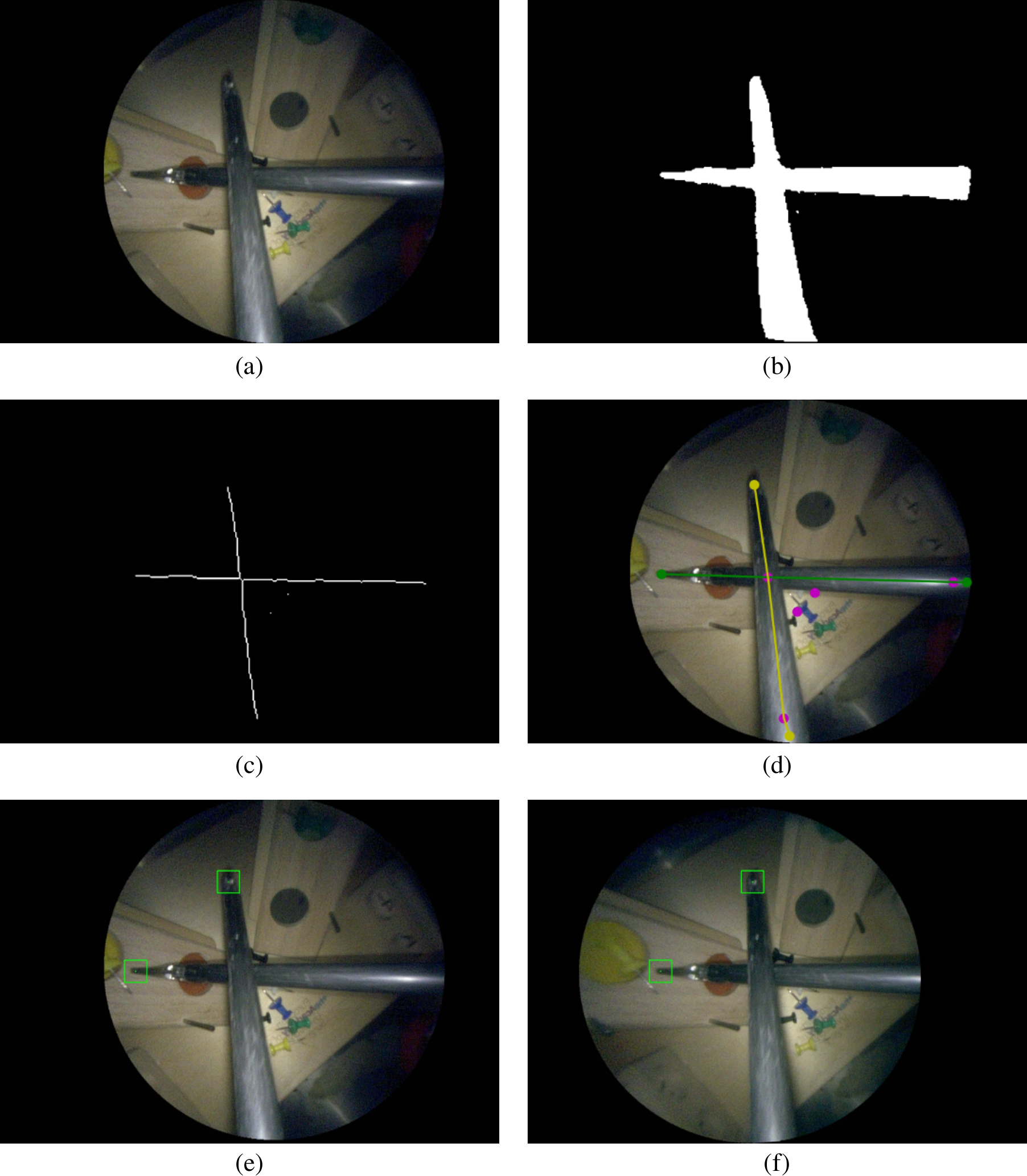}
    \caption{Instrument localization pipeline:(a) stereo-rectified right camera input image; (b) predicted tool segmentation; (c) skeletonisation; (d) graph extraction, 2D detection of entrypoints and tips, right and left instrument labelled in green and yellow; (e) left and (f) right stereo-matched tooltips in 2D (bottom row). The pink dots in (d) are graph nodes extracted from the skeleton in (c) but represent neither entrypoints nor tooltips.}
    \label{fig:instrument_localization_pipeline}
\end{figure}

A multi-step image processing pipeline was developed for markerless image-based instrument localization (see Fig. \ref{fig:instrument_localization_pipeline}). As input, the pipeline takes the raw images from a stereo endoscope. As output, it provides the 3D tip positions of the visible instruments. The maximum number of instruments and tips per instrument are required as inputs. In the task performed in our user study, presented in Sec.~\ref{sec:experiments}), a maximum of two instruments with two tips may be present. 

The 2D tooltip localization in image coordinates is a key intermediate step in this pipeline. Training a supervised bounding box detector for the tips could be a possible approach to perform the detection. However, to implement the semantically rich AIT presented in Sec.~\ref{sec:autonomous_instrument_tracking} and Fig. \ref{fig:ait_objective} we would still need to know whether the detected tips belong to the same or different instruments, and more precisely whether they belong to the instrument handled by the dominant or non-dominant hand. Therefore, we opted for estimating the more informative tool-background semantic segmentation instead. Via processing the segmentation prediction, we estimate how many instruments are in the image, localize the tips, and associate each tip with either the left or right-hand instrument.
A downside of using semantic segmentation in comparison to a detector is the increased annotation time required to build a suitable training set.
However, recent advances to reduce the number of contour annotations needed to achieve the segmentation such as \citep{Vardazaryan2018, Fuentes-Hurtado2019, Garcia-Peraza-Herrera2021} greatly mitigate this drawback.

\rebnew{In the remaining of this section we first discuss the assumptions made, imaging hardware, and preprocessing steps. Then, we proceed to describe the localization pipeline. The localization method consists of the following steps: 
binary tool-background segmentation 
(Sec. \ref{sec:instrument_segmentation}), 
skeletonization of the segmentation mask 
(Sec. \ref{sec:instrument_graph_construction}), 
graph extraction from the pixel-wide skeleton 
(Sec. \ref{sec:instrument_graph_construction}), 
entrynode detection on the graph 
(Sec. \ref{sec:instrument_entry_node_extraction}), 
leaf node detection on the graph
(Sec. \ref{sec:instrument_leaf_node_to_entry_node_matching}), 
leaf node to entry node matching
(Sec. \ref{sec:instrument_leaf_node_to_entry_node_matching}),
and left/right instrument identification
(Sec. \ref{sec:left_right_instrument_identification}).
After matching leaf nodes to entry nodes we have a subgraph for each instrument, and we distinguish between the left/right instrument using the estimated location of each instrument's entry node
(Sec. \ref{sec:left_right_instrument_identification}).
}

\rebnew{The implementation of the whole localization pipeline was done in Python, reading the video feed from the framegrabber V4L2 device with OpenCV, and performing the deep learning inference with Caffe \citep{jia2014caffe} on an NVIDIA GeForce GTX Titan X GPU.
}

\subsubsection{Assumptions of proposed instrument localization pipeline}
\label{sec:assumptions}
\begin{figure}[hb!]
    \centering
    \includegraphics[width=0.8\textwidth]{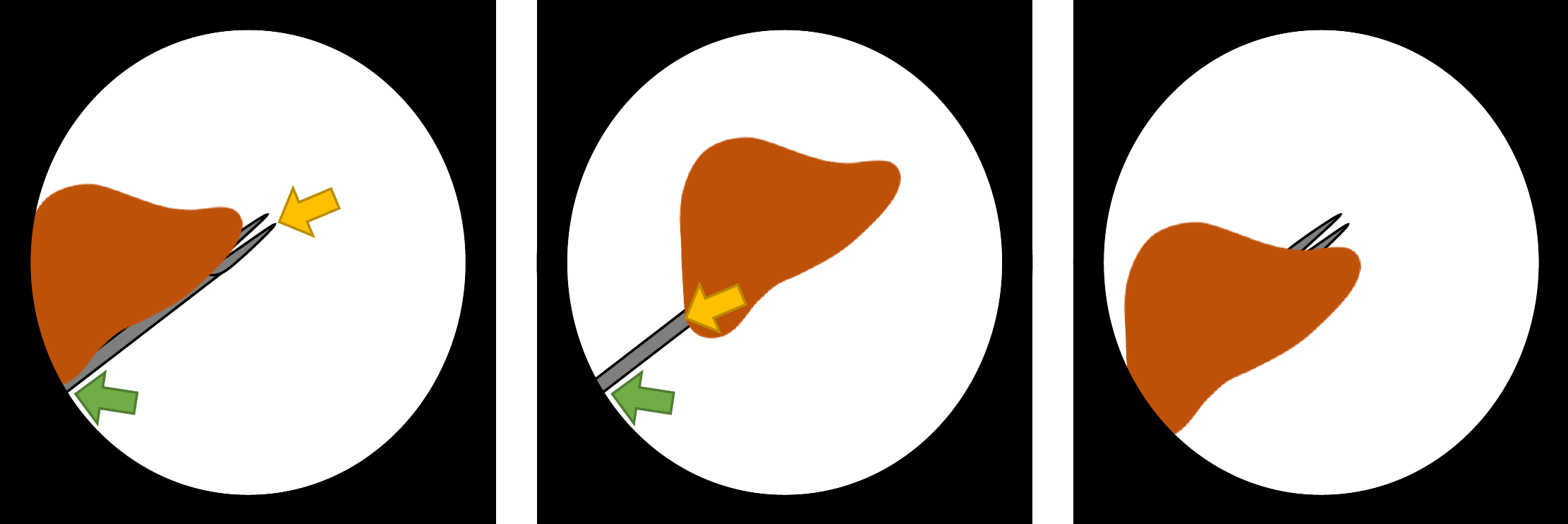}
    \caption{Behaviour and limitations of the instrument localization pipeline in the presence of occlusions. The detected entrypoint and tooltip are indicated by the green and yellow arrow, respectively. In the partially occluded instrument (left), there is a visible path from entrypoint to tip, therefore the instrument is correctly detected. However, when the tip is occluded (center), the tooltip is detected to be on the shaft. If the entrypoint is occluded (right), the instrument is not detected in this stage of the research as tools are expected to enter the scene from the boundary of the content area.}
    \label{fig:occlusion_limitation}
\end{figure}

In our instrument localization pipeline, we assume that the instruments are not completely occluded. Partial occlusions are supported, as long as there is a visible path from the \emph{entrypoint} to the tip of the instrument. Note that with \emph{entrypoint} we refer to the point located at the edge of the endoscopic content area where the instrument enters the image. This point is not to be confused with the \emph{incision point} which is the point on the body wall where the incision is made through which the instrument enters the patient's body. Now, if the tip is occluded, the tooltip will be estimated on the furthermost point of the shaft. When the entrypoint is completely covered, the instrument will not be detected in the current approach. Methods that exploit knowledge of the incision point could help in such a case (and could be explored in future work as they do not form the core of this work). The current limitations are illustrated in Fig. \ref{fig:occlusion_limitation}. The assumption that instruments have to enter from the edge serves two purposes, 1) as a noise reduction technique for the segmentation, because false positive \textit{islands} of pixels can be easily discarded, and 2) to detect whether the instrument is held by the right/left hand of 
\rebmod{the surgeon (as explained in Sec. \ref{sec:left_right_instrument_identification}).}
In most cases, the entrypoint of at least one of the instruments will be visible. Therefore, the benefits of the assumption that instruments will not be completely occluded largely outweigh its limitations. The proposal of surgical tool segmentation models that are robust to entrypoint occlusions (Fig. \ref{fig:occlusion_limitation}, right) or complete occlusions is out of the scope of this work. 

\subsubsection{Imaging hardware and preprocessing procedure}
\label{sec:imaging_hardware_and_preprocessing_procedure}
The stereo camera and endoscopy module of choice for this work were the \textsc{Tipcam1 S 3D ORL} ($30$\degree~view on a $4$\,mm outer diameter shaft) and \textsc{Image1 S D3-LINK} respectively (both from \textsc{Karl Storz}, Germany). The DVI output of the endoscopy module is plugged into a \textsc{DVI2PCIe Duo} framegrabber (\textsc{Epiphan}, Canada). The endoscopy module produces images at $60$\,fps and at a resolution of $1920\times1080$ pixels, which turns into $1920\times540$ as each grabbed image contains a stereo pair with the left frame on even rows and the right frame on odd ones.
These images are upsampled in the y-axis so that two images of $1920\times1080$ pixels are obtained. Upscaling is chosen (as opposed to downsampling to $960\times540$ pixels) to avoid degrading the depth resolution based on x-axis disparity.
\rebmod{The left-right cameras are calibrated using a chessboard pattern of $1.1$\,mm-wide squares (Cognex Glass Calibration Plate Set 320-0015R, \textsc{Applied Image Inc.}, NY, USA).
}
Both frames, left and right, are rectified in real-time. Then, the black background of the images is cropped out, keeping just a square crop of the endoscopic circle content area (as shown in Fig. \ref{fig:ait_objective}), which results in an image of $495\times495$ pixels. Finally, the image where the 2D tooltip localization is going to be performed (either the left or right frame can be chosen 
without loss of generality) is downsampled to $256\times256$ pixels to speed up the subsequent processing steps (i.e. segmentation, graph extraction and 2D tooltip localization). Once the 2D tooltips have been estimated, they are extrapolated to the original image size and the disparity estimation and 3D tooltip reconstruction in Sec.~\ref{sec:tooltip_3d_position_reconstruction} is performed on the original upsampled images of $1920\times1080$ pixels.

\subsubsection{Instrument segmentation}
\label{sec:instrument_segmentation}
In this work, we trained a CNN to segment instruments in our experimental setup (see Fig. \ref{fig:endoscope_guidance_setup}). 
While having the necessary characteristics for a bimanual laparoscopic task,
the visual appearance of the surgical training model we use is not representative of a real clinical environment.
Therefore, we do not propose a new image segmentation 
approach but rather focus on the downstream computational questions.
In order to translate our pipeline to the clinic, a newly annotated dataset containing real clinical images would need to be curated, and the images would need to contain artifacts typical of endoscopic procedures such as blood, partial occlusions, smoke, and blurring.
\rebnew{Alternatively, an existing surgical dataset could be used. 
We have compiled a list of public datasets for tool segmentation
\footnote{\url{https://github.com/luiscarlosgph/list-of-surgical-tool-datasets}
}
where the data available includes surgical scenes such as retinal microsurgery, laparoscopic adrenalectomy, pancreatic resection, neurosurgery, colorectal surgery, nephrectomy, proctocolectomy, and cholecystectomy amongst others. 
The compiled list also includes datasets for similar endoscopic tasks such as tool presence, instrument classification, tool-tissue action detection, skill assessment and workflow recognition, and laparoscopic image-to-image translation. The unlabelled data in these other datasets could also be potentially helpful for tool segmentation. 
}

\rebmod{
Next, we provide the details on how we built the segmentation model for our particular proof-of-concept of the robotic endoscope control.
The dataset we curated consists of $1110$ image-annotation pairs used for training, and $70$ image-annotation pairs employed for validation (hyperparameter tuning). 
These $1110+70$ image-annotation pairs were manually selected by the first co-authors so that the chosen images represent well the variety of scenes in the task. They have been extracted from the recording of a surgical exercise in the lab, prior to the user study, and in a different location.
There is no testing set at this point because the segmentation is an intermediary step. 
In Sec. \ref{sec:validation_of_instrument_localization_pipeline}, we give more details about our testing set, which is used to evaluate the whole tooltip localization pipeline (as opposed to just the segmentation).
The images in each stereo pair do not look the same: there is an observable difference in colour tones between them. Therefore, the data set has an even number of left and right frames such that either of them could be used as input for the surgical tool segmentation.
In the training set, $470$ images ($42$\%) do not contain any tool. In them, the endoscope just observes the task setting under different viewpoints and lighting conditions (diverse intensities of the light source). The remaining $640$ images of the training set, and all images of the validation set, have been manually labelled with delineations of the laparoscopic tools.
}
The U-Net \citep{Ronneberger2015} architecture showed superior performance in the tool segmentation EndoVis MICCAI challenge \citep{Allan2019}. Therefore, this was the architecture of choice employed for segmentation ($32$ neurons in the first layer and convolutional blocks composed of Conv + ReLU + BN). A minibatch of $4$ images is used.
Default conventional values and common practice was followed for setting the hyperparameters as detailled hereafter.
The batch normalization~\citep{Ioffe2015} momentum was set to $0.1$ (default value in PyTorch). 
Following the U-Net implementation in \citep{Ronneberger2015}, Dropout~\citep{Srivastava14a} was used. In our implementation, Dropout was employed in layers with $\geq512$ neurons ($p\!\!=\!\!0.5$), as in \citep{Garcia-Peraza-Herrera2017}. 
Following \cite{Bengio2012}, the initial learning rate (LR) of choice was set to $1e-2$.
The network was trained for a maximum of $100$ epochs. As is common practice, LR decay was employed during training, multiplying the LR by $0.5$ every $10$ epochs.
Data augmentation was limited to on-the-fly left-right flips.
As we evaluate our segmentation using the intersection over union (IoU), our loss function \rebnew{$\mathcal{L}_{\textrm{IoU}}$} is a continuous approximation to the intersection over union \citep{Rahman2016a} averaged over classes:
\begin{equation}
\begin{split}
    \label{eq:miou_loss}
I(\mathbf{\hat{y}}, \mathbf{y}, k) &= \sum_{i=1}^{P}\hat{y}_{i,k} \cdot y_{i, k}, \\
U(\mathbf{\hat{y}}, \mathbf{y}, k) &= \sum_{i=1}^{P}\hat{y}_{i,k} + \sum_{i=1}^{P} y_{i, k} - \sum_{i=1}^{P}\hat{y}_{i,k} \cdot y_{i, k}, \\
\mathcal{L}_{\textrm{IoU}}(\mathbf{\hat{y}}, \mathbf{y})
        &= 1 - \frac{1}{K} \sum_{k=1}^{K} 
\frac{I(\mathbf{\hat{y}}, \mathbf{y}, k) + \epsilon}
{U(\mathbf{\hat{y}}, \mathbf{y}, k) + \epsilon},
\end{split}
\end{equation}
where $P$ is the number of pixels, $K=2$ is the number of classes (instrument and background), $\mathbf{\hat{y}}$ represents the estimated probability maps, 
\rebmod{$\mathbf{y}$ represents the ground truth probability maps,
$\hat{y}_{i,k}$ is the estimated probability of the pixel $i$ belonging to the class $k$, and $y_{i, k}$ is the ground truth probability of the pixel $i$ belonging to class $k$.
}
A machine epsilon $\epsilon$ is added to prevent divisions by zero (e.g., in case that both prediction and ground truth are all background). 

Once we have obtained a segmentation prediction from the trained convolutional model, we proceed to convert the segmentation into a graph, which is a stepping stone towards the tooltip detection.

\subsubsection{Instrument graph construction}
\label{sec:instrument_graph_construction}
The instrument segmentation prediction is skeletonized via medial surface axis thinning \citep{Lee1994a}. 
The resulting skeleton is converted via the Image-Py skeleton network framework \citep{Xiaolong} into a pixel skeleton graph $G = (V, E)$ (see Fig. \ref{fig:entrypoint}e), where $V$ is a set of vertices and 
$E \subseteq \{ \{x, y]\} : x, y \in V \land x \ne y \}$ is a set of edges. The nodes $v_i \in V$ are defined as a tuple $v_i = (i, \boldsymbol{p_i})$ where $i$ and  $\boldsymbol{p_i} = \{x_i, y_i\}$ represent the node index and $2D$ point image coordinates, respectively.

\subsubsection{Instrument entry node extraction}
\label{sec:instrument_entry_node_extraction}
\begin{figure}[tb]
    \centering
    \includegraphics[width=0.60\textwidth]{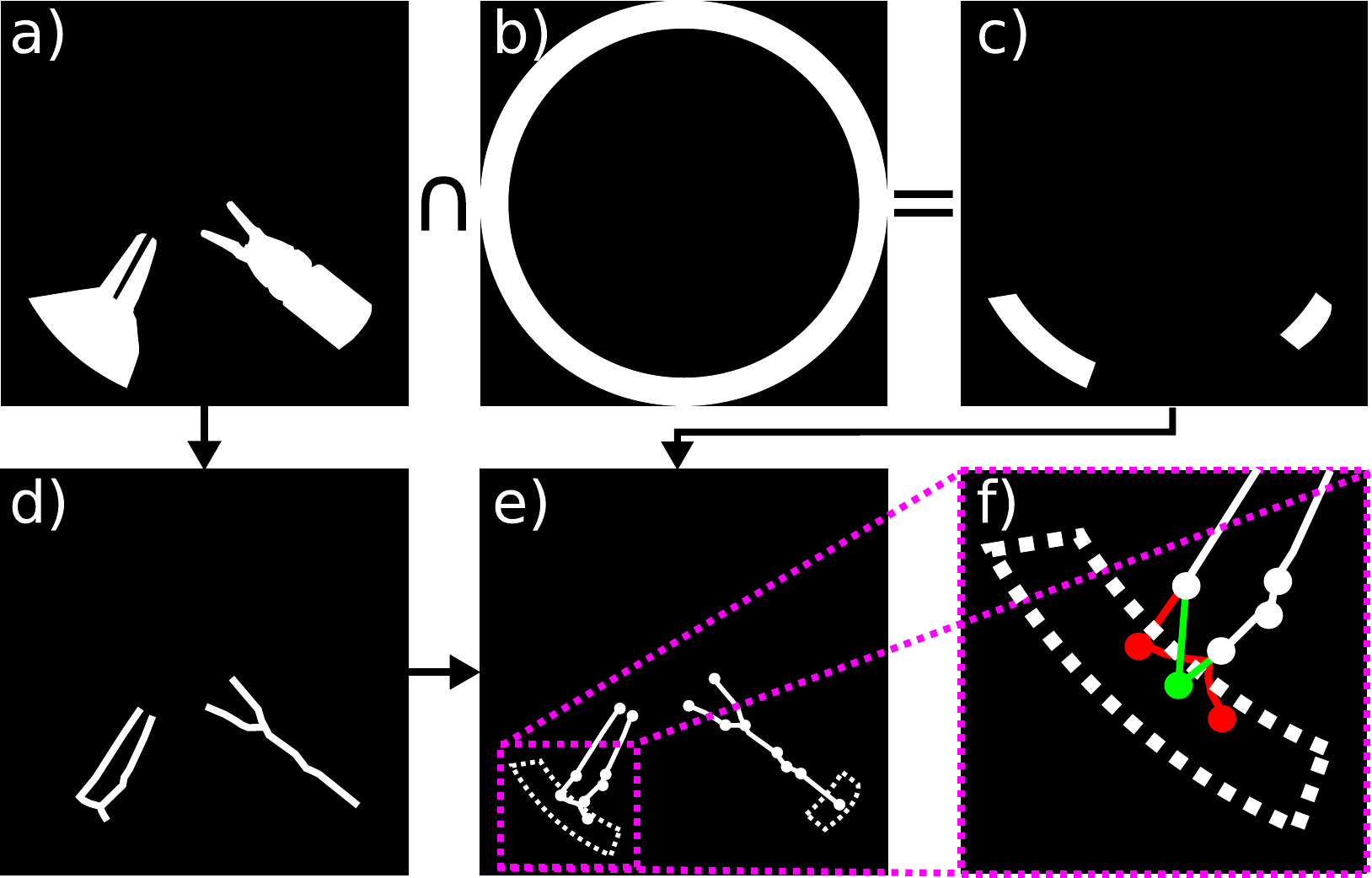}
    \caption{Surgical instrument graph construction and entry node extraction: a) segmentation mask; b) circle mask used to detect entrypoints; c) intersection of segmentation mask and circle mask; d) segmentation mask skeleton obtained according to \citep{Lee1994a}; e) graph obtained from skeleton by means of \citep{Xiaolong}; f) entrypoint detection. If several graph nodes lie inside the entrypoint mask (in red), they are merged into a new single entry node (in green) whose position attribute is set to the centroid of all the graph nodes inside the dotted area. 
}
    \label{fig:entrypoint}
\end{figure}

As the size of the image is known, a circular segmentation mask (see Fig.\ref{fig:entrypoint}b) is used to detect the graph nodes that could potentially correspond to instrument entrypoints. That is, given $G$, we populate a set $R$ containing those graph nodes that represent tool entrypoints into the endoscopic image.
Those graph nodes contained within the intersection of the circle mask and the tool segmentation mask are collapsed into a single new \textit{entry} node $v_{c} = (n, \boldsymbol{p_c})$ per instrument, where $\boldsymbol{p_c} = \{x_c, y_c\}$ is set to the centroid of all nodes captured within the aforementioned intersection. See Fig. \ref{fig:entrypoint}b-\ref{fig:entrypoint}f for an example of entry node extraction. 

A depth-first search is launched from each entry node to determine all the graph nodes that can be reached from entry nodes. Those that cannot be reached are pruned from the graph.

\subsubsection{Instrument leaf node to entry node matching}
\label{sec:instrument_leaf_node_to_entry_node_matching}
Let $L = \{v \in V : d_{G}(v) = 1 \land v \not\in R\}$ be the set containing all \textit{leaf} nodes, where $d_G(v) = | \{ u \in V : \{u, v\} \in E \} |$. In this part of the instrument localization pipeline each \textit{leaf} node in $L$ is paired to an entrypoint node in $R$. This is solved by recursively traversing $G$, starting from each \textit{leaf}. The criteria to decide which node to traverse next is coined in this work as \emph{dot product recursive traversal}. It is based on the assumption that the correct path from a tip to a corresponding entrypoint is the one with minimal direction changes. The stopping condition is reaching an entry node.

\begin{figure}[tb]
    \centering
    \includegraphics[width=0.42\textwidth]{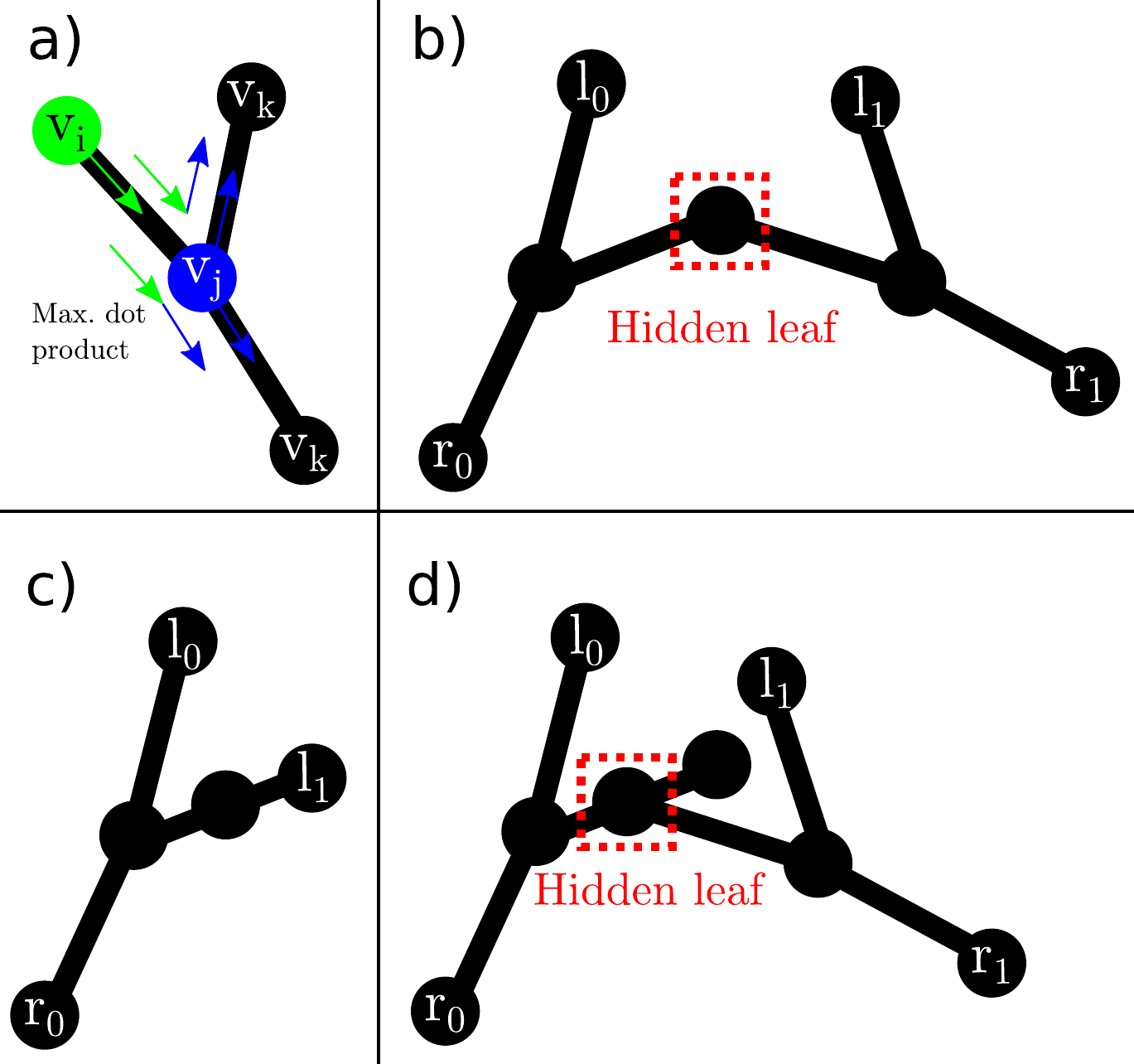}
    \caption{Leaf traversal and hidden leaves: a) graph traversal step. $v_i$ (green) represents the previous node. $v_j$ (blue) is the current node. $v_k$ are possible next nodes. Following \eqref{eq:dot_criteria}, the next node will be the one that maximizes the dot product; b) possible \textit{hidden leaf} (dashed red box) connected to two nodes; c) node with two neighbours that does not represent a \textit{hidden leaf} as both connecting edges are labelled after dot product traversal from $l_1$; d) possible \textit{hidden leaf} (dashed red box) with three neighbours.}
    \label{fig:cases}
\end{figure}

Dot product recursive traversal operates as follows. Let $v_i, v_j \in V$ be two arbitrary nodes, and $\{v_i, v_j\}$ the undirected edge connecting them. Assuming $v_i$ is previously visited and $v_j$ being traversed, the next node $v*$ to visit is chosen following:
\begin{equation}
\begin{split}
    \label{eq:dot_criteria}
    v^* = \argmax_{(k, \boldsymbol{p_k}) \in N(v_j) - A} 
        \left( 
            \frac{\boldsymbol{p_j} - \boldsymbol{p_i}}{\vert\vert \boldsymbol{p_j} - \boldsymbol{p_i} \vert\vert_2} \cdot \frac{\boldsymbol{p_k} - \boldsymbol{p_j}}{\vert\vert \boldsymbol{p_k} - \boldsymbol{p_j} \vert\vert_2}
        \right),
\end{split}
\end{equation}
where $N(v_i)\!=\!\{ w\!\in\!V\!:\!\{v_i, w\} \!\in\!E\}$, and $A\!=\!\{v_i\}$ is the set of nodes previously traversed. The idea behind \eqref{eq:dot_criteria} is to perform a greedy minimization of direction changes along the path from tooltip to entrypoint (Fig. \ref{fig:cases}a).

In the case of two tools present in the image, it is possible to have \textit{hidden leaves} (Fig.~\ref{fig:cases}b and \ref{fig:cases}d), defined as graph nodes that represent an overlap between the tip of an instrument and another instrument. This situation can easily occur (Fig. \ref{fig:cases}) in surgical tasks, including the task presented in the experiments from Sec.~$\ref{sec:experiments}$. There are two possible graph arrangements that can lead to \textit{hidden leaves}. A node with exactly two (Fig. \ref{fig:cases}b) or three neighbours (Fig. \ref{fig:cases}d). Nonetheless, the number of neighbours alone does not facilitate the discovery of such \textit{hidden leaves} (and subsequent disentanglement of tools), as it is also possible for a node with exactly two (could be a chain instead) or three (could be a bifurcation instead) neighbours to not be a \textit{hidden leaf} (see Fig. \ref{fig:cases}c). Hence, extra information is needed. Aiming in this direction, after each successful traversal from a normal \textit{leaf} to an entry node, all the edges along the path are labelled with the index of the entry node. In addition, all the edges directly connected to an entry node are also labelled. 

A node with exactly two or three neighbours whose edges are all labelled with different entry nodes 
is a \textit{hidden leaf}.
Labelling helps to solve some of the \textit{hidden leaf} cases.
Such leaves can be duplicated, effectively splitting the graph into two, and disentangling the overlapped instruments. After disentanglement, they become normal leaves which can be assigned to an entry node by dot product traversal \eqref{eq:dot_criteria}. Although not a \textit{hidden leaf}, a node with exactly four neighbours whose edges are labelled represents an overlap which can be trivially disentangled.
\textit{Hidden leaves} such as the ones presented in Fig. \ref{fig:cases}b and \ref{fig:cases}d cannot be classified with certainty as such just with the graph/skeleton information. As shown in Fig. \ref{fig:cases}, different tool configurations/occlusions could lead to the same graph configuration. As not all the tips can be unambiguously detected, entry nodes that are unmatched after dot product traversal (i.e., they were not reached after launching a traversal from each leaf node to a possible entry node) are paired to the furthest opposing node connected to them.

Although the traversal from tips to entrypoints has been illustrated in this section with one or two instruments (as it is the case in our setup, see Fig. \ref{fig:endoscope_guidance_setup}), the dot product traversal generalizes to setups with more instruments as the assumption that the path from tip to entrypoint is the one with less direction changes still holds.

\subsubsection{Instrument graph pruning}
\label{sec:instrument_graph_pruning}
Noisy skeletons can lead to inaccurate graphs containing more than two leaves matched to the same entry node, or more than two entry nodes connected to leaves. 
In our framework, unmatched entry nodes are deleted. In addition, due to the nature of our experimental setup, a maximum of two tools with two tips each can be found. Therefore, when more than two leaves are matched to the same entry node, only the two furthest are kept. 
Analogously, when more than two entry nodes are found and matched to leaves, the two kept are those with the longest chain (from entry node to leaves).
That is, a maximum of two disentangled instrument graphs remain after pruning. 

\subsubsection{Left/right instrument identification}
\label{sec:left_right_instrument_identification}

\begin{figure}[hb!]
    \centering
    \includegraphics[width=0.60\textwidth]{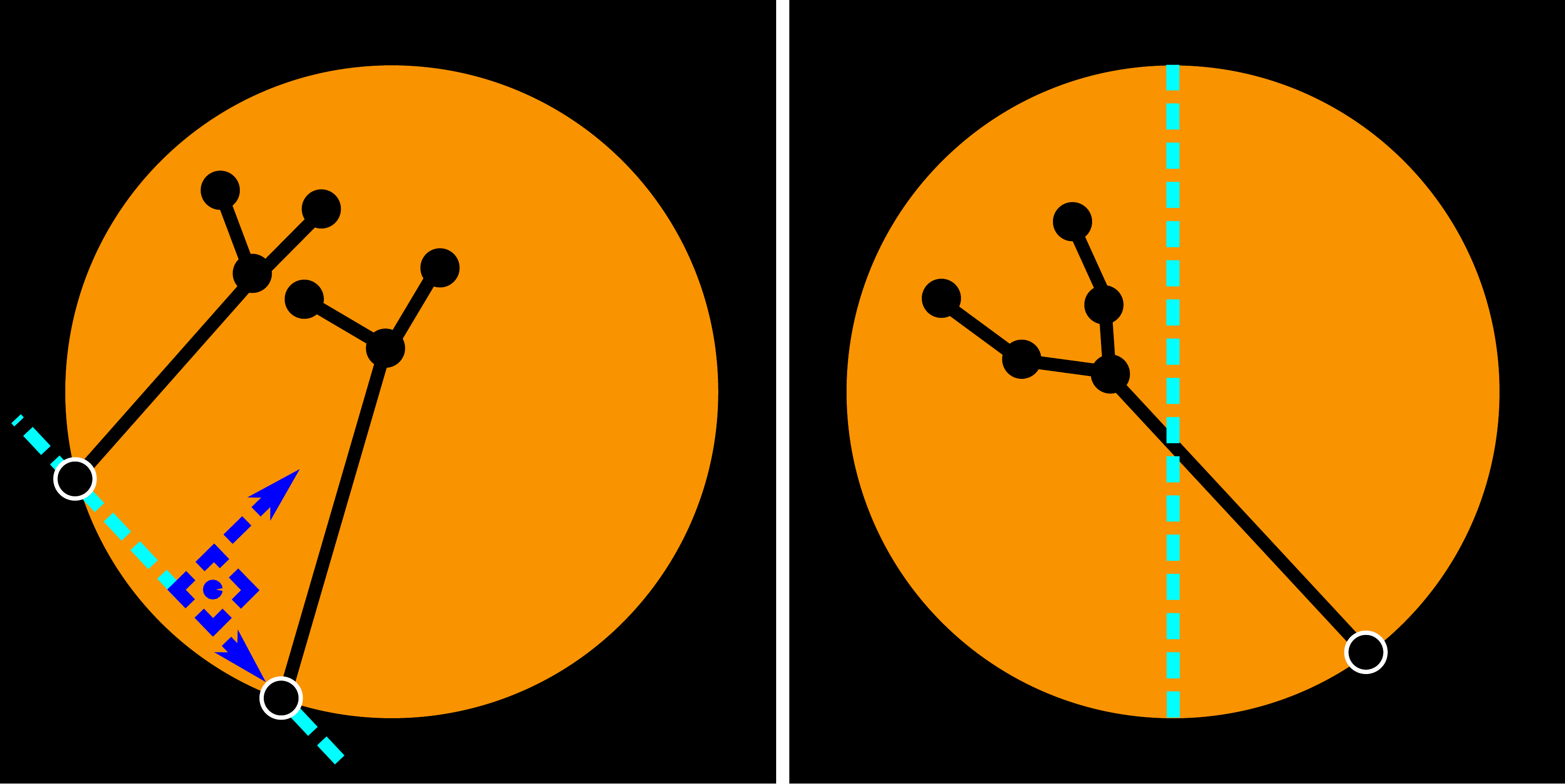}
    \caption{Left/right instrument identification. Two-instrument case (left). Single-instrument case (right). The location of the entrypoints is used to identify whether the instruments are left/right. 
When two instruments are visible, an imaginary vertical line (parallel to the vertical axis of the image) that crosses over the central point of the segment connecting both entrypoints is used to determine if the instrument is left/right. 
When there is only one instrument, the location of the entrypoint with regards to the vertical axis of the image determines which tool is visible. If the entrypoint resides in the right half, as in the figure above, this is considered to be the right instrument.
}
    \label{fig:left_right}
\end{figure}

In the presence of a single instrument, left/right vertical semi-circles determine whether the instrument is left/right (see Fig. \ref{fig:left_right}, right), i.e. if the entrypoint of the tool is located in the right half of the image, it is assumed that the subgraph connected to this entrypoint is the right instrument, and viceversa. Note that this simple method is also generalizable to scenarios with three to five instruments, which are different from the two-instrument solo surgery setting examined in this work (see Fig. \ref{fig:autonomous_lastt_model}), but still worth mentioning as there are some endoscopic procedures that may involve such number of tools \citep{Abdi2016}.

When two instruments are detected (i.e. two entrypoints with their corresponding subgraphs), a line segment connecting the entrypoints of both instruments is assumed to be the viewing horizon. A vertical line that is parallel to the vertical axis of the image and cuts through the central point of the viewing horizon defines whether the entrypoints (and subsequently the instruments) are left/right (see Fig. \ref{fig:left_right}, left).

\subsubsection{Tooltip stereo matching}
\label{sec:tooltip_3d_position_reconstruction}
Once the tips of the instruments have been detected and classified as right/left instrument, the disparity for each tooltip in the left and right stereo images is estimated using classical intensity-based template matching. As endoscope images are stereo-rectified, template matching with a sliding window of $64\times64$ pixels running over the same line (and only in one direction) suffices for the stereo matching. Normalized cross-correlation is the cost function of choice. Given the disparity measure for each tooltip, its depth can be reconstructed using the pinhole camera model and conventional epipolar geometry. The 3D reconstruction was performed with an extended Kalman filter (EKF). The EKF is valuable here, because of its capacity to bridge potential measurement gaps and to reduce the noise on the 3D position estimate, which is very sensitive to small disparity variations, as the lens separation of the \textsc{Tipcam} is only 1.6\,mm. The details of the EKF are specified in Sec.~\ref{sec:ekf}.

Although in our proposed experimental setup we use stereovision because we have an stereo-endoscope, many centers still use monoscopic endoscopes. In this case, a method such as that presented by Liu et al. \citep{Liu2020} could be used to estimate the 3D tip location directly from the 2D endoscopy. \section{Visual servoing for robotic endoscope control}
\label{sec:visual_servoing}

A visual servoing controller determines the relative motion between a camera and an observed target in order to produce the desired camera view upon the target. In the case of AIT, the target is the (virtual) instrument tip position $\vec{s}$, defined by \eqref{eq:virtual_instrument_tip}, and the camera is the endoscope held by the robotic endoscope holder. When working with endoscopes, the visual servoing controller needs to take into account a number of aspects specific for endoscopy, including the presence of the incision point which imposes a geometric constraint and the endoscope geometry. For the online estimation of the incision point, automated routines exist, such as \citep{Gruijthuijsen2018RAL, Dong2016ICRA}. This section formalizes visual servoing approaches for REC in MIS.
 
\subsection{Visual servoing approaches}

Two classical approaches exist for visual servoing problems: image-based visual servoing (IBVS) and position-based visual servoing (PBVS)~\citep{Chaumette2008}. 
For REC, an extension to these methods is necessary as the camera motion is constrained by the presence of the incision point. In IBVS, this can be done by modifying the interaction matrix, such that it incorporates the kinematic constraint of the incision point~\citep{Osa2010} or such that it describes the mapping between the image space and the joint space of the robotic endoscope holder~\citep{Uecker1995,Zhao2014}. As these IBVS approaches only act in the image plane, the zoom level can be controlled by a decoupled depth controller~\citep{Chen2018}. PBVS approaches can incorporate the incision constraint in an inverse kinematics algorithm that computes the desired robot pose, given the desired endoscopic view~\citep{Yu2013,Eslamian2016}.

Implementations of the above approaches, that the authors are aware of, lack generality: they are formulated for a specific robotic endoscope holder and do not cover oblique-viewing endoscopes, while such endoscopes are commonly used in MIS procedures. Yet, oblique-viewing endoscopes are the most challenging to handle for clinicians \citep{Pandya2014}, and could thus reap most benefits of REC. Generic constraint-based control frameworks, such as eTaSL \citep{etasl2014}, could be applied with a generalized endoscope model, like presented below, although they are slower than explicit visual servoing methods. 

\subsection{Visual servoing with generalized endoscope model}

This section introduces a novel generalized endoscope model for visual servoing that incorporates the incision constraint, as well as the endoscope geometry. Such a model is presented here, along with the ensuing modifications to the classical IBVS and PBVS approaches.

\subsubsection{Generalized endoscope model}
\label{sec:generalized_endoscope_model}

Endoscopes come in different forms and sizes. Rigid endoscopes are typically straight, but can also be pre-bent. The camera can be oriented collinear with the longitudinal axis of the scope or can be positioned at an oblique angle. Some scopes are flexible over their entire length, others contain a proximal rigid straight portion with a distal bendable portion. 

\begin{figure}[hb!]
    \centering
    \includegraphics[width=0.45\textwidth]{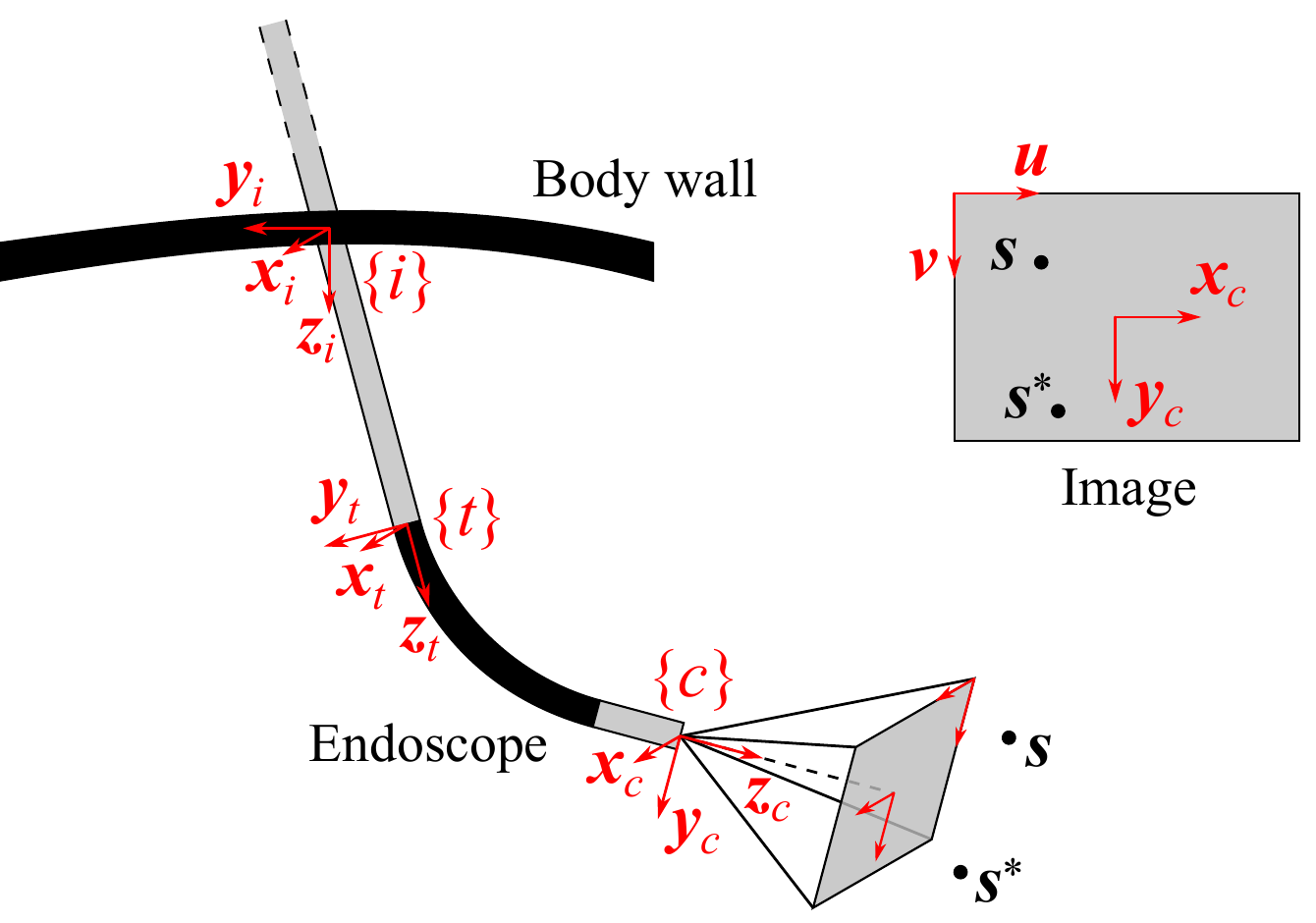}
    \caption{Definition of reference frames in a generalized endoscope model. The incision frame $\{i\}$ is located at the incision point in the body wall, the distal tip frame $\{t\}$ at the end of the straight endoscope shaft and the camera frame $\{c\}$ at the end of the endoscope, in the optical center of the camera. The image produced by the endoscope is also shown (upper right insert), along with the projections of the detected feature of interest $\vec{s}$ and its desired position $\vec{s}^*$, and with the image coordinate vectors $(\vec{u},\vec{v})$.}
    \label{fig:ref_frames}
\end{figure}

\begin{figure}[hb!]
    \centering
    \includegraphics[width=0.45\textwidth]{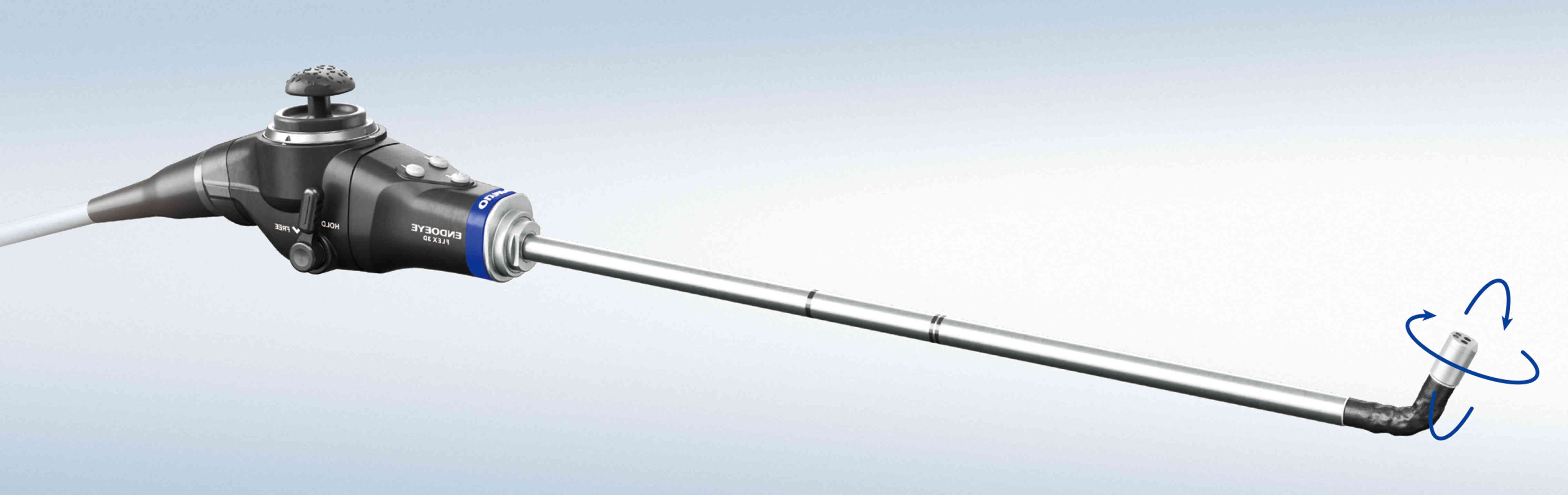}
    \caption{\textsc{EndoEye Flex 3D} \copyright \textsc{Olympus Corporation} (Tokyo, Japan).}
    \label{fig:endoeye}
\end{figure}

Fig. ~\ref{fig:ref_frames} visualizes a general endoscope geometry that encompasses all the above configurations, along with the frames of interest. The incision frame $\{i\}$ is defined at the incision point and is common for all robotic endoscope holders. The $z$-axis of $\{i\}$ is the inward-pointing normal of the body wall. A frame $\{t\}$ is connected to the distal tip of the straight portion of the endoscope shaft, with its $z$-axis pointing along the shaft axis. In the most general case, the camera frame $\{c\}$ is located at an offset and rotated with respect to $\{t\}$. The offset can reproduce for tip articulation, for stereo camera lens separation. The rotation can account for oblique-viewing endoscopes. As such, this endoscope model can describe complex endoscopes, such as the articulating 3D video endoscope \textsc{EndoEye Flex 3D} (\textsc{Olympus}, Japan) (Fig.~\ref{fig:endoeye}).

Starting from this general endoscope model, different visual servoing approaches for REC will be detailed next. The visual servoing approaches strive to determine the endoscope motion that is needed to match the position $\vec{s}=\left[\begin{matrix}x & y & z\end{matrix}\right]^T$ of the feature of interest, expressed in the camera frame, with its desired position $\vec{s}^*=\left[\begin{matrix}x^* & y^* & z^*\end{matrix}\right]^T$, while taking into account the presence of the incision point. The visual servoing approaches assume that the robot endoscope holder has three active DoFs that can produce any linear velocity of the endoscope tip. In order to obtain a fully determined endoscope motion, it is further assumed that the remaining rolling DoF about the endoscope axis is not controlled by the visual servoing controller, but by an external controller. Note that, as was pointed out in Sec.~\ref{sec:autonomous_instrument_tracking},  this DoF could be employed to control the camera horizon.

The following notation will be used in the subsequent sections: a rotation of angle $\xi$ about the axis $i$ will be denoted by $\mat{R}_i(\xi)$. For a transformation from a frame $\{j\}$ to a frame $\{i\}$, the notation $\mat{T}_j^i$ will be used, consisting of a rotation $\mat{R}_j^i$ and a translation $\mat{p}_j^i$. Further, the twist vector $\vec{t}=\begin{bmatrix}\vec{v}^T & \vec{\omega}^T\end{bmatrix}^T$ is defined as the concatenation of a linear velocity $\vec{v}$ and an angular velocity $\vec{\omega}$. For all kinematic variables, the reference frames will be indicated with a trailing superscript. For the features $\vec{s}$ and the error $\vec{e}$ in the camera frame, the trailing superscript $^c$ is mostly omitted for brevity.

\subsubsection{EKF for tooltip 3D position reconstruction}
\label{sec:ekf}
The instrument localization pipeline from Sec.~\ref{sec:instrument_localization_pipeline} yields the tooltip image coordinates $u_l, v_l$ and the disparity $d_x$. The 3D tooltip position, required for the visual servoing methods, is estimated from these measurement data, through an EKF. The state transition model describes a linear tooltip motion of exponentially decreasing velocity, partially expressed in frames $\{i\}$ and $\{c\}$ to limit the non-linearity, and the observation model implements the pinhole camera model:
\begin{align}
\vec{x}_k &= g(\vec{x}_{k-1},\vec{t}^c_{c,k}) + \vec{\epsilon}_k,\nonumber\\
&= \begin{bmatrix}
\vec{s}^c_{k-1} + \Delta T (\mat{R}^c_i \dot{\vec{s}}^i_{k-1} + \mat{L}_{3D}(\vec{s}^c_{k-1}) \vec{t}^c_{c,k}) \\
\lambda_s \dot{\vec{s}}^i_{k-1}
\end{bmatrix} + \vec{\epsilon}_k,
\\
\vec{z}_k &= h(\vec{x}_{k}) + \vec{\delta}_k
= \begin{bmatrix}
\frac{x_k f_x}{z_k} + c_x \\
\frac{y_k f_y}{z_k} + c_y \\
\frac{b_c f_x}{z_k}
\end{bmatrix} + \vec{\delta}_k,
\end{align}
where 
$\vec{x}_k = {\begin{bmatrix} \vec{s}_k^{c^T} & {\dot{\vec{s}}}_k^{i^T}\end{bmatrix}}^T$ is the state vector, 
$\vec{s}_k^c$ is the tooltip position in the camera frame $\{c\}$, $\dot{\vec{s}}^i_k$ is the tooltip velocity in the incision frame $\{i\}$, $\vec{t}^c_{c}$ is the camera twist,
$\mat{L}_{3D}$ is the 3D interaction matrix from  Eq.~\eqref{eq:l3d},
$\lambda_s$ is a reduction factor $<1$ (governing the exponential decrease of $\dot{\vec{s}}^i$),
$\vec{z}_k = \begin{bmatrix} u_{l,k} & v_{l,k} & d_{x,k}\end{bmatrix}^T$ is the observation vector,
$f_x, f_y, c_x, c_y$ are the intrinsic camera parameters, $b_c$ the distance between the optical centres of the (left and right) cameras,
and $\vec{\epsilon}_k$ and $\vec{\delta}_k$ are the usual process and observation noises. The velocity reduction factor $\lambda_s$ is introduced to scale down the contribution of dead reckoning during measurement gaps.

\subsubsection{Image-based visual servoing (IBVS)}
\label{sec:ibvs}

IBVS aims to determine the camera motion to move the 2D projection of the 3D feature point $\vec{s}$ to its desired position in the image plane. Assuming a pinhole camera model, the 2D projection $\vec{s}_n$ is obtained by expressing $\vec{s}$ in normalized camera coordinates:
\begin{equation}
    \vec{s}_n = \left[\begin{matrix}x_n & y_n\end{matrix}\right]^T = \left[\begin{matrix} x/z & y/z\end{matrix}\right]^T.
\end{equation}
Classically, the relation between the camera twist $\vec{t}^c_c$ and the 2D feature point velocity $\dot{\vec{s}}_n$ is expressed by the interaction matrix $\mat{L}_{2D}$~\citep{Chaumette2008}:
\begin{equation}
    \dot{\vec{s}}_n = \mat{L}_{2D} \vec{t}^c_c,
    \label{eq:sndot}
\end{equation}
where
\begin{equation}
    \mat{L}_{2D} = \left[\begin{matrix}
    -1/z & 0 & x_n/z & x_n y_n & -(1+x_n^2) & y_n \\
    0 & -1/z & y_n/z & 1+y_n^2 & -x_n y_n & -x_n \\
    \end{matrix}\right].
    \label{eq:l2d}
\end{equation}
In this equation, it is assumed that the camera has six DoFs, while only three are available for the endoscope control. To incorporate these constraints, the camera twist $\vec{t}^c_c$ needs to be mapped first to the twist of tip of the endoscope's straight portion $\vec{t}^t_t$: 
\begin{equation}
    \vec{t}^c_c = \mat{J}^c_t \vec{t}^t_t, 
\end{equation}
where $\mat{J}^c_t$ is the well-known expression for a twist transformation:
\begin{equation}
    \mat{J}^c_t = \left[\begin{matrix}
    \mat{R}^c_t & -\mat{R}^c_t [\vec{p}^t_c]_{\times} \\
    \vec{0} & \mat{R}^c_t \\
    \end{matrix}\right].
    \label{eq:jct}
\end{equation}
The operator $[\,]_{\times}$ is the operator for the skew-symmetric matrix. The incision constraint introduces a coupling between the linear tip velocity $\vec{v}^t_t$ and angular tip velocity $\vec{\omega}^t_t$ and can be expressed as:
\begin{equation}
    \vec{t}^t_t = \mat{J}_i \vec{v}^t_t,
\end{equation}
with the incision transformation
\begin{equation}
    \mat{J}_i = \left[\begin{matrix}
    1 & 0 & 0 \\
    0 & 1 & 0 \\
    0 & 0 & 1 \\
    0 & -1/l & 0 \\
    1/l & 0 & 0 \\
    0 & 0 & 0 \\
    \end{matrix}\right],
    \label{eq:ji}
\end{equation}
and the inserted endoscope length $l=\|\vec{p}^i_t\|$. Combining \eqref{eq:sndot}-\eqref{eq:ji} yields the modified interaction matrix $\mat{L}^\prime_{2D}$:
\begin{equation}
\mat{L}^\prime_{2D} = \mat{L}_{2D} \mat{J}^c_t \mat{J}_i
\end{equation}
which maps $\vec{v}^t_t$ to $\dot{\vec{s}}_n$.
This matrix is a generalized form for the modified interaction matrix presented in~\citep{Osa2010}.

As is customary in visual servoing, the error in the normalized image space is expressed as 
\begin{equation}
    \vec{e}_n = \vec{s}_n - \vec{s}^*_n
\end{equation}
and the control law enforces an exponential decay of the error:
\begin{equation}
    \dot{\vec{e}}_n = -\lambda \vec{e}_n,
\end{equation}
characterized by the time constant $\tau = 1 / \lambda$. For a constant $\vec{s}^*_n$, this yields the desired endoscope tip velocity:
\begin{align}
    \label{eq:en}
    \dot{\vec{e}}_n &= \dot{\vec{s}}_n = \mat{L}^\prime_{2D} \vec{v}^t_t = -\lambda \vec{e}_n \nonumber \\
    &\Rightarrow \vec{v}^t_t = -\lambda \mat{L}^{\prime+}_{2D} \vec{e}_n.
\end{align}

\subsubsection{Image-based visual servoing with decoupled depth control (IBVS+DC)}
\label{sec:dcvs}

IBVS only seeks to optimize the 2D projected position $\vec{s}_n$ of the  target point $\vec{s}$ in the image plane.
As such IBVS alone is insufficient to control the 3D position of the endoscope. A decoupled depth controller can be added to control the third DoF. This was proposed in~\citep{Chen2018} and will be generalized here. 

The depth controller acts along the $z$-axis of the camera frame $\{c\}$ and uses the kinematic relation between the camera twist $\vec{t}^c_c$ and the change in the depth $z$ of $\vec{s}$:
\begin{equation}
    \dot{z} = \mat{L}_z\vec{t}^c_c,
\end{equation}
where
\begin{equation}
    \mat{L}_z = \left[\begin{matrix}0 & 0 & -1 & -y & x & 0\end{matrix}\right].
\end{equation}

To reduce the depth error $e_z = z - z^*$, concurrently with the image-space error $\vec{e}_n$, a similar reasoning as with IBVS can be followed, yielding:
\begin{align}
    \left[\begin{matrix}\dot{\vec{e}}_n \\\dot{e}_z\end{matrix}\right]
    &= \left[\begin{matrix}\dot{\vec{s}}_n \\\dot{z}\end{matrix}\right]
    = \left[\begin{matrix}\mat{L}_{2D} \\\mat{L}_z\end{matrix}\right] \mat{J}^c_t \mat{J}_i \vec{v}^t_t
    = \left[\begin{matrix}\mat{L}^\prime_{2D} \\\mat{L}^\prime_z\end{matrix}\right] \vec{v}^t_t
    = -\lambda \left[\begin{matrix}\vec{e}_n \\ e_z\end{matrix}\right] \nonumber \\
    &\Rightarrow \vec{v}^t_t = -\lambda \left[\begin{matrix}\mat{L}^\prime_{2D} \\\mat{L}^\prime_z\end{matrix}\right]^+ \left[\begin{matrix}\vec{e}_n \\ e_z\end{matrix}\right].
\end{align}
To differentiate between directions, it is possible to define $\lambda$ as a diagonal matrix, rather than as a scalar.

\subsubsection{3D image-based visual servoing (3D IBVS)}
\label{sec:3dibvs}

Instead of decoupling the control in the image plane and the depth control, the 3D feature $\vec{s}$ can also be used directly to define the 3D motion of the endoscope. This requires a 3D interaction matrix $\mat{L}_{3D}$, which can be derived from the kinematic equations of motion for the stationary 3D point $\vec{s}$ in the moving camera frame $\{c\}$:
\begin{equation}
    \dot{\vec{s}} = -\vec{v}^c_c - \vec{\omega}^c_c \times \vec{s} = \mat{L}_{3D} \vec{t}^c_c,
    \label{eq:l3d}
\end{equation}
with
\begin{equation}
    \mat{L}_{3D} = \left[\begin{matrix}-\mat{I} && [\vec{s}]_{\times}\end{matrix}\right].
\end{equation}
As before, the modified interaction matrix $\mat{L}^{\prime}_{3D}$ can be obtained by including the offset of the tip frame with respect to the camera frame and the incision constraint. The desired endoscope velocity that ensures an exponential decay of the error $\vec{e} = \vec{s} - \vec{s}^*$ follows then from:
\begin{align}
    \dot{\vec{e}} &= \dot{\vec{s}} = \mat{L}_{3D} \mat{J}^c_t \mat{J}_i \vec{v}^t_t = \mat{L}^\prime_{3D} \vec{v}^t_t = -\lambda \vec{e} \nonumber \\
    &\Rightarrow \vec{v}^t_t = -\lambda \mat{L}^{\prime+}_{3D} \vec{e}.
\end{align}

\subsubsection{Position-based visual servoing (PBVS)}
\label{sec:pbvs}

PBVS identifies the camera pose, with respect to an external reference frame, that produces the desired view upon the 3D feature $\vec{s}$ and moves the camera towards this pose. As mentioned before, the camera pose is constrained to three DoFs due to the presence of the incision point and the separate horizon stabilization. Finding the desired camera pose, while taking into account its kinematic constraints, involves solving the inverse kinematics for the endoscope as defined in Fig.~\ref{fig:ref_frames}. 

The forward kinematics of the endoscope can be described as a function of three joint variables ($\theta_1$, $\theta_2$, $l$). Based on these variables, any endoscope pose can be reached by applying successive operations in a forward kinematics chains. When these joint variables are set to zero, the endoscope system is in a configuration where the incision frame $\{i\}$ coincides with the distal tip frame $\{t\}$, while the camera frame is offset by $\vec{p}_c^t$ and rotated by $\mat{R}_c^t=\mat{R}_x(\alpha)$, with $\alpha$ the oblique viewing angle of the endoscope. Starting from this configuration, $\theta_1$ rotates $\{t\}$ about its $y$-axis, then $\theta_2$ rotates it about its $x$-axis and finally $l$ translates it along its $z$-axis. This leads to the following forward kinematic equations, expressed in the reference frame $\{i\}$: 
\begin{align}
    \mat{T}^i_c \tilde{\vec{s}} &= {\mat{T}^i_c}^* \tilde{\vec{s}}^* \nonumber \\
    \label{eq:inv_kin}
    &= {\mat{T}^i_t}^* \mat{T}^t_c \tilde{\vec{s}}^* \\
&= \left[\begin{matrix}
    \mat{R}_y(\theta^*_1) \mat{R}_x(\theta^*_2) & \vec{0} \\
    \vec{0}^T & 1
    \end{matrix}\right]
    \left[\begin{matrix}
    \mat{I} & l^* \hat{\vec{e}}_3 \\
    \vec{0}^T & 1
    \end{matrix}\right]
    \left[\begin{matrix}
    \mat{R}_x(\alpha) & \vec{p}^t_c \\
    \vec{0}^T & 1
    \end{matrix}\right] 
    \tilde{\vec{s}}^*, \nonumber
\end{align}
with $\hat{\vec{e}}_3=\left[\begin{matrix}0 & 0 & 1\end{matrix}\right]^T$ the unit vector along the $z$-direction. The trailing $^*$ designates a desired value, different from the current value. The $\tilde{\ }$ signifies the homogeneous representation of a 3D vector. Equation~\eqref{eq:inv_kin} constitutes a system of three equations in the unknowns $(\theta^*_1,\theta^*_2,l^*)$. \ref{app:inverse_kinematics} elaborates the analytic solution to this inverse kinematics problem.

The solution of the inverse kinematics can be inserted in the forward kinematics equations to obtain the desired position of the distal endoscope tip ${\vec{p}^i_t}^*$:
\begin{equation}
    {\vec{p}^i_t}^* = \left[\begin{matrix}
    l^* \sin(\theta^*_1) \cos(\theta^*_2) \\
    -l^* \sin(\theta^*_2) \\
    l^* \cos(\theta^*_1) \cos(\theta^*_2)
    \end{matrix}\right],
\end{equation}
which straightforwardly leads to the position error of the distal tip, expressed with respect to the incision frame $\{i\}$:
\begin{equation}
    \vec{e}^i = \vec{p}^i_t - {\vec{p}^i_t}^*.
\end{equation}
When an exponential decaying error is required, the desired endoscope velocity becomes:
\begin{align}
    \vec{v}^i_t &= \dot{\vec{e}}^i = -\lambda \vec{e}^i \\
\end{align}
and can be expressed in the frame $\{t\}$ as:
\begin{align}
    \vec{v}^t_t = -\lambda \mat{R}^t_i \vec{e}^i.
\end{align}

\subsection{Simulation of visual servoing methods}
\label{sec:vs_discussion}

\begin{figure}[tb]
    \centering
    \includegraphics[width=0.42\textwidth]{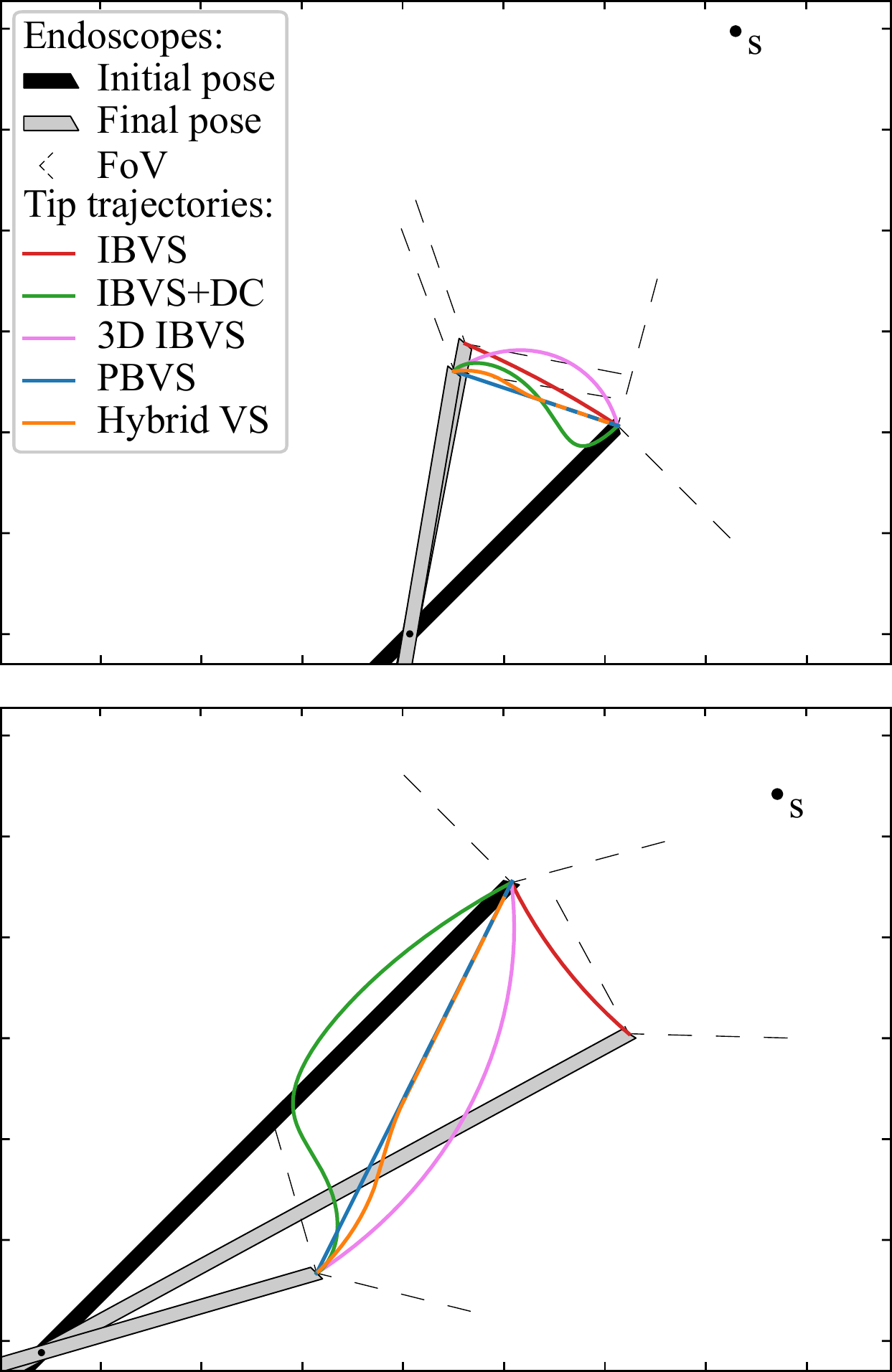}
    \caption{Comparison of the endoscope tip trajectories for different visual servoing approaches for REC. In this simulation, an oblique-viewing $30\degree$ endoscope with a 120$\degree$ FOV was used. Its initial and final pose are drawn. In the final pose, the feature of interest $\vec{s}$ is in the desired position with respect to the camera. The trajectories are drawn for a case with a small initial depth error (top) and a large initial depth error (bottom).}
    \label{fig:vs_methods}
\end{figure}

A simulation was implemented to validate all four visual servoing methods for REC: IBVS, IBVS+DC, 3D IBVS and PBVS. Fig.~\ref{fig:vs_methods} presents a visual comparison between them, for a $30\degree$ oblique-viewing endoscope with a $120\degree$ FoV. In all simulations, $\vec{s}$ enters the FoV from a side. The visual serviong controller moves the endoscope to center $\vec{s}$ within its FoV at a given depth $z^*$, or $\vec{s}^*=\left[\begin{matrix}0 & 0 & z^*\end{matrix}\right]^T$. The trajectories described by the endoscope tip are shown in the graphs, as well as the initial (marked in black) and final (marked in grey) endoscope poses.

\subsubsection{Comparison of visual servoing methods}

From the graphs, it is clear that IBVS differs from the other approaches in that, by construction, it does not attain the desired depth $z^*$ in the final endoscope pose. Moreover, IBVS also doesn't guarantee a constant depth $z$. Consequently, $z$ will drift towards undesired depths over time. In some configurations, this can be counteracted by separately controlling $l$ to stay constant, but this does not hold for the general case. IBVS alone is thus unsuitable for REC and 3D information about $\vec{s}$ is a requirement.

Both IBVS+DC and 3D IBVS linearize the visual servoing problem in the camera frame. This enables a desired exponential decay of the targeted errors, but does not produce a well-controlled endoscope motion in Cartesian space. It can be seen from Fig.~\ref{fig:vs_methods} that the trajectories for these methods deviate from the straight trajectory that is accomplished by PBVS, and more so for large initial errors. As space is limited in REC and the environment delicate, the straight trajectory of PBVS appears favourable compared to its alternatives.

IBVS typically yields more accurate visual servoing results than PBVS, because the feedback loop in IBVS-based methods can mitigate camera calibration errors (excluding stereo calibration errors). However, the objective in REC often is to keep $\vec{s}$ inside a specific region of the endoscopic image (cf. position hysteresis), rather than at an exact image coordinate. The importance of the higher accuracy of IBVS is thus tempered by this region-based control objective: small calibration inaccuracies are acceptable. Therefore, and in contrast to the claims in \citep{Osa2010}, it can be argued that the predictability of a straight visual servoing trajectory outweighs the importance of the visual servoing accuracy. This argumentation points out why PBVS is the preferred approach for REC, especially when large initial errors exist.

\subsubsection{Hybrid PBVS and 3D IBVS}
\label{sec:hybrid_vs}

If the accuracy of PBVS would need to be enhanced, e.g., when significant calibration errors exist, it is possible to apply a hybrid visual servoing method. PBVS can be used until the initial error drops below a certain threshold and from there, the visual servoing controller gradually switches to an IBVS-based approach for refinement, by applying a weighted combination of the desired tip velocities $\vec{v}^t_t$ computed by each visual servoing method. The curved shape of IBVS trajectories can thus be suppressed. In experiments that are  not further documented here, it was observed that 3D IBVS, which ascertains an exponential Cartesian error decay, provided a more predictable and thus more desirable endoscope behaviour than IBVS+DC. To ensure robustness against potential calibration errors, the hybrid combination of PBVS and 3D IBVS was thus selected for the experiments in Sec.~\ref{sec:experiments}. Fig.~\ref{fig:vs_methods} shows the simulated performance of the hybrid visual servoing approach, which gradually transitions from PBVS to 3D IBVS when the error $\|\vec{e}_n\|$ in the normalized image space goes from $0.6$ to $0.3$. \section{Experiments}
\label{sec:experiments}
\begin{figure}[hb!]
	\centering
	\includegraphics[width=0.7\columnwidth]{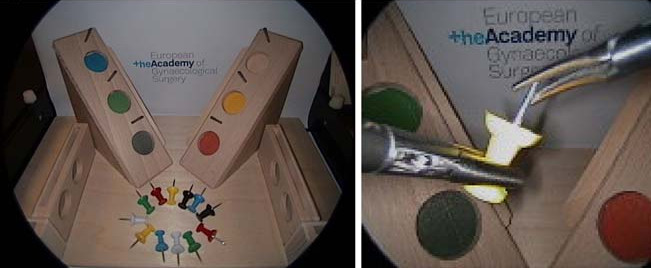}
	\caption{LASTT \citep{LASTT2020} laparoscopic training model. Inital position of the pins before starting the bi-manual coordination task (left). Procedure to pass a pin from the non-dominant to the dominant hand (right).}
	\label{fig:autonomous_lastt_model}
\end{figure}

\begin{figure}[hb!]
    \centering
    \includegraphics[width=0.7\columnwidth]{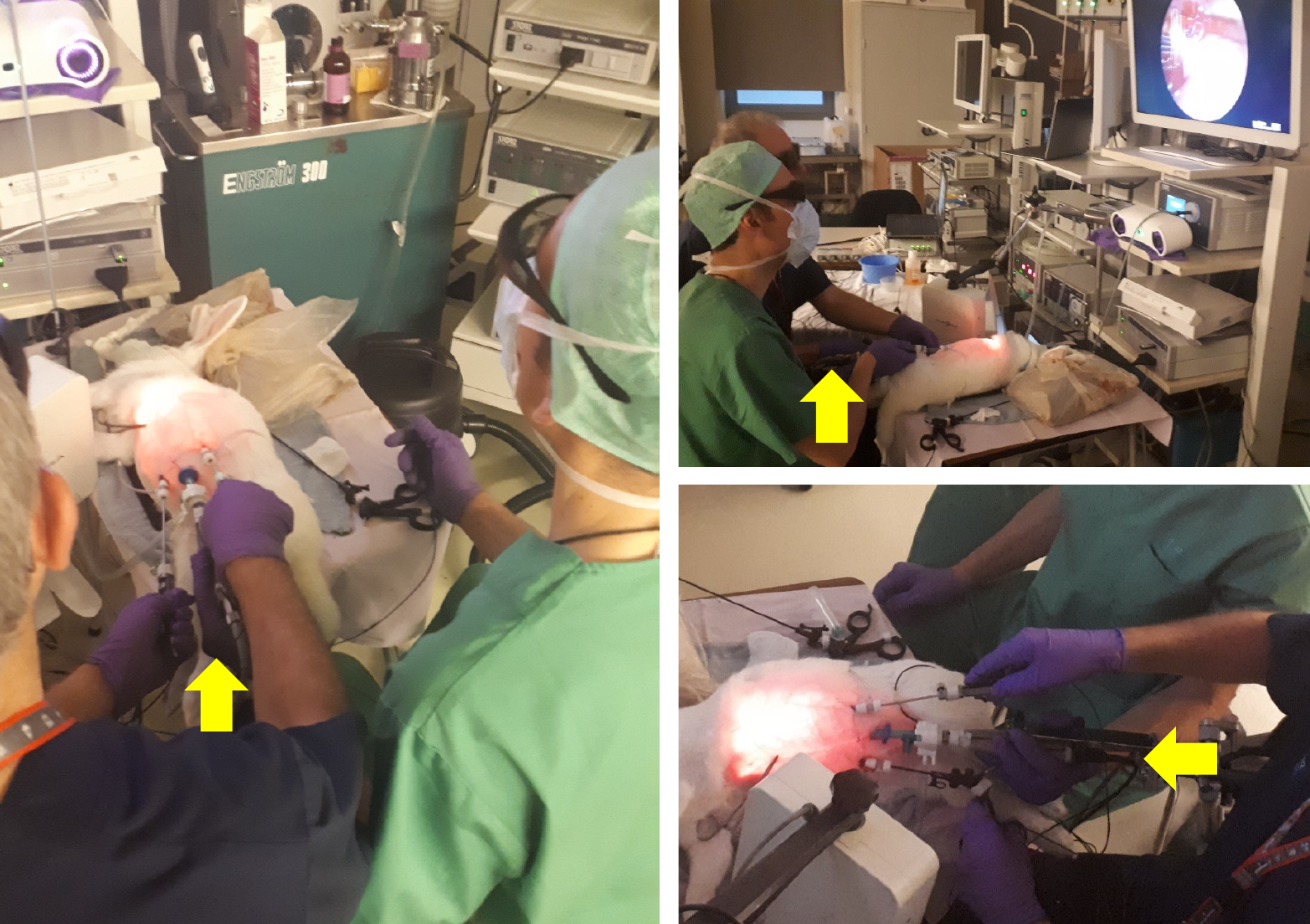}
    \caption{Spina bifida intervention performed on an animal model. The surgeon dressed in blue scrubs controls the instruments and manipulates the tissue. The colleague dressed in green guides and holds the endoscope camera during the intervention. The yellow arrows point to the hand of the assistant guiding the camera. As becomes evident in the pictures above, this operating arrangement is not ergonomic, leading to discomfort that increases with the duration of the intervention, and severely limiting the tasks that the surgeon controlling the camera can perform. Picture courtesy of Prof. Jan Deprest.}
    \label{fig:fetal_inspiration}
\end{figure}

To determine the feasibility of the proposed autonomous endoscopy framework, an experimental setup was built (see Fig.~\ref{fig:endoscope_guidance_setup}). The mockup surgical setting consisted of a laparoscopic skills testing and training model (LASTT) placed within a laparoscopic box trainer (see Fig.~\ref{fig:autonomous_lastt_model}). A \emph{bi-manual coordination} exercise was chosen as the target surgical task for the experiments. In this task, a set of pushpins need to be passed between hands and placed in the right pockets. 
The choice of both laparoscopic trainer and surgical task was clinically motivated. The present study is largely inspired by the surgical scenario occurring during spina bifida \cite{Bruner1999a,Meuli2014,Kabagambe2018} surgical procedures (see Fig. \ref{fig:fetal_inspiration}). In this fetal treatment, a surgeon operates while another one guides the endoscope. 
The LASTT model along with the bi-manual coordination task have been developed by The European Academy for Gynaecological Surgery~\footnote{\url{https://esge.org/centre/the-european-academy-of-gynaecological-surgery}} as an initiative to improve quality control, training and education in gynaecological surgery \citep{Campo2012}. Therefore, they are ideal candidates for the feasibility study of the proposed autonomous endoscopy framework.

\subsection{Bi-manual coordination task}
\label{sec:bi_manual_coordination_task}
This task starts by placing a set of coloured pushpins at the base of the LASTT model (see Fig.~\ref{fig:autonomous_lastt_model}, left). There are two pins of each colour. The operator has to pick a pin with the non-dominant hand, pass it to the dominant hand (Fig.~\ref{fig:autonomous_lastt_model}, right), and place it inside the pocket of the same colour. The LASTT task is successfully completed when a pushpin of each of the six colours has been placed in a corresponding pocket, within less than five minutes. If the pin is dropped during the procedure, a second pin of the same colour has to be picked up from the base. If the second pin is also dropped, the test is considered a failure. 

\rebnew{As shown in the demonstration video of the bi-manual coordination task with the LASTT model of the European Academy of Gynaecological Surgery\footnote{
    \url{https://europeanacademy.org/training-tools/lastt/}
},
this exercise cannot be performed with a fixed immobile endoscope due to the reduced field of view of the endoscope, the limited space available for maneuvers within the operating cavity, and the small size of the pins (which resembles small tissue structures). All of these characteristics of the LASTT model mimic the real operating conditions, particularly for gynaecological interventions. 
}
Without a robotic endoscope holder, the bi-manual coordination task is performed with one trainee handling the laparoscopic graspers and another trainee acting as the (human) camera assistant. The assistant should hold the endoscope and keep the view centered on what the laparoscopic operator is doing. 
\rebmod{In our experiments, this human camera assistant is replaced by the \textsc{Virtuose6D}\footnote{https://www.haption.com/en/products-en/virtuose-6d-en.html} (\textsc{Haption SA}, Laval, France) robotic arm.
As shown in \citep{Avellino2020}, the dimensions, workspace, and supported payload of this robotic arm are well suited for robotic endoscope control
\footnote{\url{https://www.youtube.com/watch?v=R1qwKAWFOIk}
}.
The operational workspace is defined as a cube of side $450\,$mm and is located in the center of the workspace envelope. The extremities of the workspace envelope are bounded by a volume of $1330\times575\times1020$\,mm$^3$.
The payload supported by the \textsc{Virtuose6D} is $35\,$N (peak)/ $10\,$N (continuous). Additionally, the Virtuose6D features passive gravity compensation, which can be mechanically adjusted to carry up to $8\,$N. Therefore, although in our setup we are using a stereo-endoscope, this system is also able to hold laparoscopy cameras (e.g. those used in abdominal surgery).
In our setup, the robotic arm was holding the \textsc{Karl Storz} \textsc{Tipcam1}, as shown in Fig.~\ref{fig:endoscope_guidance_setup}. The \textsc{Virtuose6D} was programmed to respond to semantically rich AIT instructions (Sec.~\ref{sec:autonomous_instrument_tracking}). Additionally, it featured a comanipulation fallback mode (Sec.~\ref{sec:comanipulation fallback}), which it naturally supports owing to its mechanical backdrivability. 
}

\subsection{Study participants}
A total of eight subjects participated in the study. Two \textit{surgeons}, two \textit{plateau novices}, and four \textit{novices}. The \textit{plateau novices} were authors of the study, who started out as novices, but familiarized themselves with the system and the task until they reached a plateau in the learning curve. Each participant performed the bi-manual coordination task five times. Before these trials, each participant practised 5--10 minutes to perform the task while assisted by the robotic endoscope holder.

\subsection{Configuration of the autonomous endoscope for the study}
The autonomous endoscope controller implemented the \emph{Hybrid PBVS and 3D IBVS} method (Sec.~\ref{sec:hybrid_vs}), switching from PBVS to 3D IBVS when the error $\|\vec{e}_n\|$ in the normalised image space decreased from $0.6$ to $0.3$. 
The target position of the endoscope tip was set to $\vec{s}^*=\begin{bmatrix}0 & 0 & z^*\end{bmatrix}^T$, where $z^* = 8$\,cm. The endoscope tip was controlled to track a trajectory towards its desired position with the tip velocity $\vec{v}^t_t$ limited to $2$\,cm/s.
This trajectory was implemented as a soft virtual fixture, with stiffness of $0.3$\,N/mm. The aforementioned low speed and stiffness were found to provide smooth and predictable motions.
\rebnew{They proved also helpful in avoiding sudden motions when one of the instruments is occluded and the remaining one is located in the violation zone.
}
Low speed and stiffness were also necessary because of the $340$\,ms delay on the measurements updates of $\vec{s}$. The framegrabber was responsible for $230$\,ms of this delay. A framegrabber that supports NVIDIA GPUDirect, not available to us at the time of writing, could be used to mitigate this latency. The other $110$\,ms came from the 3D tooltip localisation pipeline (Sec.~\ref{sec:instrument_localization_pipeline}). A delay that could be potentially reduced in future work using TensorRT. Measurements were available at $9$\,Hz.

The position hysteresis approach, which was illustrated in Fig.~\ref{fig:ait_objective}, was applied separately in the image plane and along the viewing direction. In the image plane, the target zone A occupied the first $40$\% of the endoscopic image radius, the transition zone B the next $20$\%, and the violation zone C the remaining $40$\%. Along the viewing axis, the target zone was set to $3$\,cm in both directions of $z^*$. The violation zone started at a distance of $5$\,cm with respect to $z^*$. The EKF for stereo reconstruction (Sec.~\ref{sec:tooltip_3d_position_reconstruction}) was used to fill missing data up to $1$\,s after the last received sample. When the instruments were lost from the view for more than $10$ seconds, the REC switched from the AIT mode to the comanipulation fallback mode, waiting to be manually reset to a safe home position.

\rebmod{When designing the experiments, two preliminary observations were made: (1) the AIT instruction that fixes the tracking target on the tip of the instrument held by the dominant hand was most convenient, and (2) instructions to change the zoom level were not used. The latter observation is easily explained by the nature of the LASTT task, which requires overview rather than close-up inspections. The former observation points out that it is confusing to track a virtual instrument tip in between the real tooltips. While concentrated on the task, participants tend to forget their non-dominant hand and move it out of the view (or in and out of the view without a particular reason). This affected the position of the virtual instrument tip in unexpected ways. In those situations where tracking the non-dominant hand is relevant (e.g., when passing the pin from one hand to another), participants quickly learned to keep the tips together. Hence, tracking the dominant-hand tool was sufficient to provide a comfortable view, and this became the only operation mode that participants used. In fact, as an operator, it was convenient to know that the system is tracking your dominant hand: this is easy to understand and remember. Thus, during all the experiments, the only instruction that was issued was to make the camera track the dominant hand. As all participants were right-handed, $\vec{s}$ was assigned to the tip of the right-hand tool.
}

 \section{Results and discussion}
\label{sec:results_and_discussion}

\rebnew{In this section we provide quantitative results on the tooltip tracking accuracy, the responsiveness and usability of the visual servoing endoscopic guidance, and the learning curve of the user study participants.
}

\subsection{Validation of instrument localization pipeline}
\label{sec:validation_of_instrument_localization_pipeline}

Given an endoscopic video frame as input, the tooltip localization pipeline produces an estimate of the 2D location of the surgical instrument tips (see Fig. \ref{fig:instrument_localization_pipeline}). The tooltip location in image coordinates is  used for the later 3D position reconstruction of the tooltips. Therefore, we first validate the localization pipeline performance independently of the overall task. This includes the instrument segmentation (Sec.~\ref{sec:instrument_segmentation}) together with the subsequent tooltip detection steps (Sec.~\ref{sec:instrument_graph_construction}--\ref{sec:left_right_instrument_identification}). 

\begin{figure}[hb!]
    \centering
    \includegraphics[width=0.42\columnwidth]{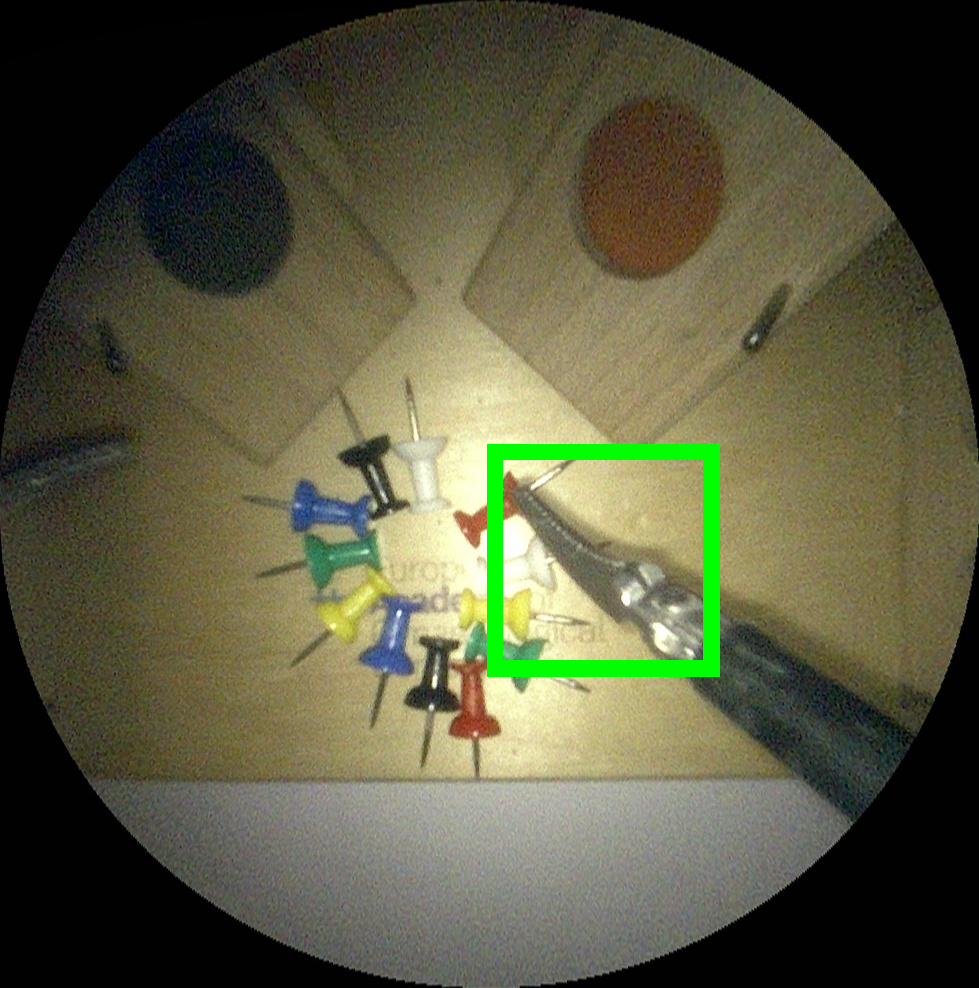}
    \caption{Image crop of visible area of 1080p endoscope video frame. The green square is the  200$\boldsymbol{\times}$200 pixel bounding box used to evaluate detection performance. An intersection over union $\boldsymbol{\geq50\%}$ between predicted and ground truth bounding boxes is considered a correct detection.}
    \label{fig:bbox}
\end{figure}

For the selected bi-manual coordination task of Sec.~\ref{sec:experiments}, two laparoscopic instruments are used. Hence, a maximum of four tips may be encountered in any given endoscopic video frame. Two for the left and two for the right instrument. We define a bounding box around each detected tooltip. The chosen size for the bounding box is $200\times200$ pixels (cf. $1080$p raw video frames). This corresponds to the size of the instrument distal part at a practical operation depth (Fig. \ref{fig:bbox}). A comparison between the bounding box and the image size is shown in Fig. \ref{fig:bbox}. 

Following common practice in object detection \citep{Everingham2014}, a $\geq50\%$ intersection over union (IoU) between the prediction and ground truth bounding boxes is considered a true positive. A predicted bounding box that does not surpass this threshold represents a false positive. The Hungarian method is employed to match predictions to ground truth bounding boxes. The number of unmatched or missed bounding boxes from the ground truth represents the false negatives. In object detection, precision and recall at different confidence levels are commonly blended into a single performance metric, the average precision~(AP) \citep{Everingham2014}. In the absence of a confidence level, we report precision and recall. 

\rebmod{The testing set that we use to report results for the whole tooltip localization pipeline comprises $379$ images.
These images are evenly sampled video frames extracted at a constant frequency from the recording of the user study experiments, when participants operate the robot (i.e. they are not used during the training or validation of the segmentation model).
}
\rebmod{Our tooltip tracking localization pipeline achieved a tooltip detection precision and recall of 72.45\% and 61.89\%, respectively. In 84.46\% of the video frames, at least one of the present tips was correctly detected.
}

\subsection{Responsiveness of the endoscopic guidance}
\begin{figure}[tb]
	\centering
	\includegraphics[width=\columnwidth]{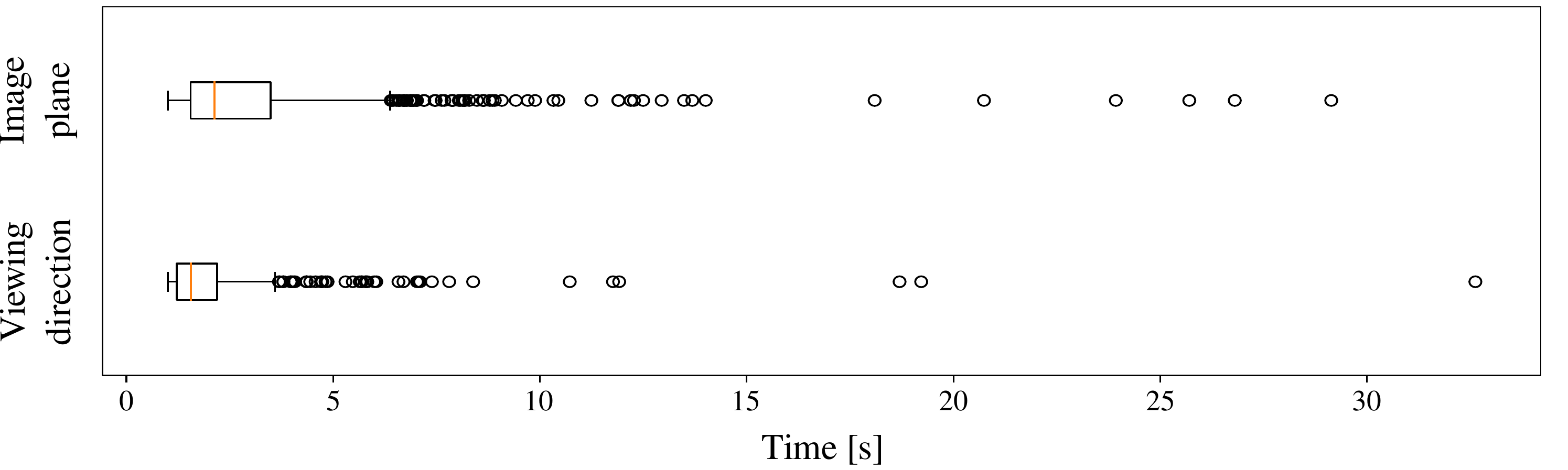}
	\caption{Time taken to correct the position of the endoscope after the dominant-hand tooltip entered the violation zone.}
	\label{fig:autonomous_correction_time}
\end{figure}

\rebmod{The responsiveness of the proposed system was also evaluated.
To navigate outside the view, participants have to place the tip of the instrument in the violation zone C (see Fig. \ref{fig:ait_objective} for a description of the zones). When this occurs, the AIT functionality is triggered until the instrument appears in zone A. Fig.~\ref{fig:autonomous_correction_time} shows how long it took for the system to recover (entering zone A) after a violation (entering zone C) was detected. As shown in the figure, the control was responsive, taking an average of $\approx3$s ($\approx2$s in the viewing direction) to bring back the instrument tips to zone A. 
The slight difference between the correction time in the viewing direction and image plane is due to the difference in size of the zone A,
which was relatively large along the viewing direction and therefore harder to violate. 
}

\rebmod{In Fig.~\ref{fig:autonomous_correction_time}, a number of outliers are present. This occurred when the participants moved their hands too fast for the REC to follow, causing the instruments to entirely disappear from the FoV. On most of these occasions, the participants were able to put the instruments back inside FoV after some time, resuming normal navigation. However, in three instances (of the outliers $>10$s), the endoscope had to be manually brought back to a safe, centered home position, using its comanipulation fallback mode.
}

\subsection{Usability of the endoscopic guidance}
\begin{figure}[htb!]
	\centering
	\includegraphics[width=\columnwidth]{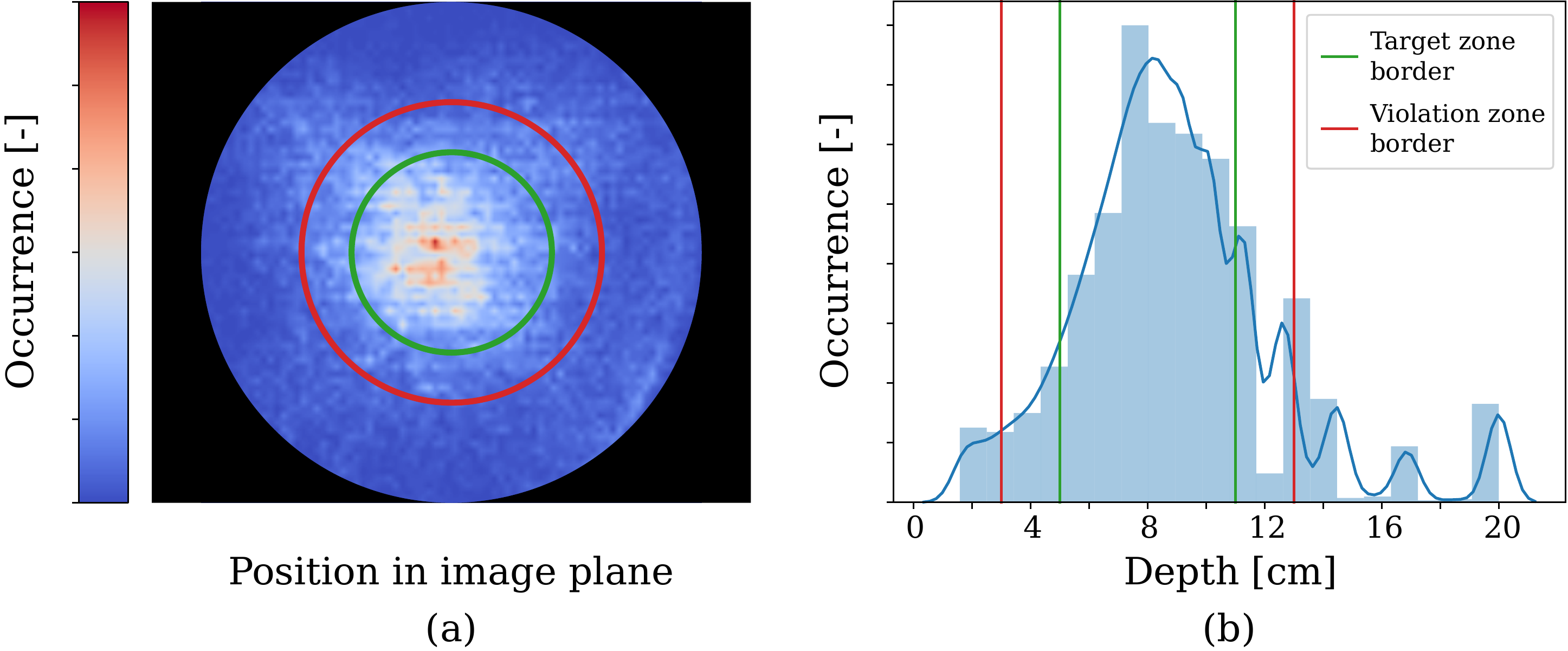}
	\caption{Distribution of dominant-hand tooltip presence across all the experiments in the 2D image and viewing direction.}
	\label{fig:autonomous_right_tip_distribution}
\end{figure}

\rebmod{When a human trainee is operating the endoscope, it is important for the coordination and the overview of the surgeon that the view remains centred around the instrument.
This is also the objective when a human trainee is operating the endoscope. To quantify this aspect, 
Fig.~\ref{fig:autonomous_right_tip_distribution} shows the distribution of tip positions for the dominant-hand instrument across all the experiments. The REC indeed manages to keep the tooltip within the boundaries of the target zone A for most of the time. In the 2D image plane, the tip of the instrument was $46\%$, $23\%$, and $31\%$ in target, transition, and violation zones, respectively. Similar behaviour was observed along the viewing direction, with a cumulative zone presence of $66\%$, $22\%$, and $12\%$, respectively.
}

\subsection{Surgical skills assessment and learning curve on the bi-manual coordination task}
\begin{figure*}[hb!]
	\centering
	\includegraphics[width=.9\textwidth]{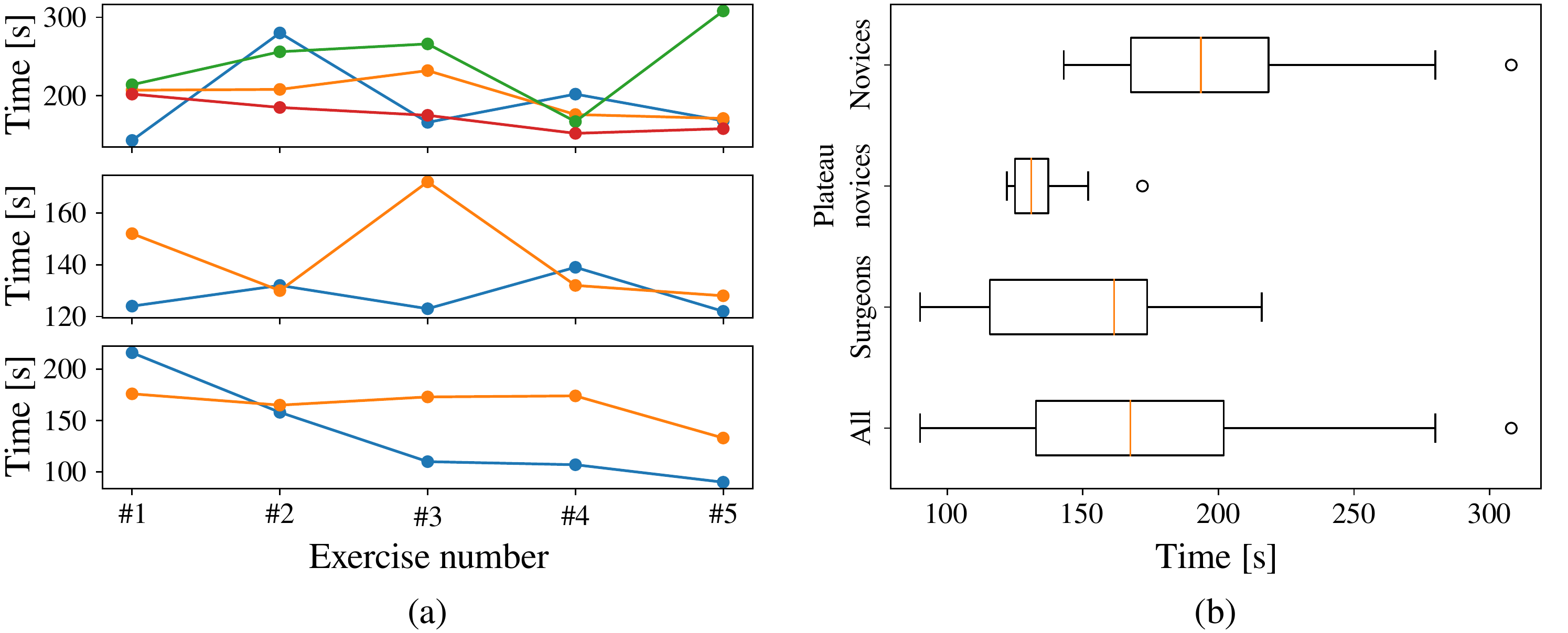}
	\caption{Completion time for all the participants in the bi-manual coordination task. (a) Completion time across attempts, with \textit{novices} (top), \textit{plateau novices} (centre), and \textit{surgeons} (bottom). (b) Completion time per group across all trials.}
	\label{fig:autonomous_lastt_completion_time}
\end{figure*}

\rebmod{The proposed system allowed the user study participants to perform the benchmark surgical task \footnote{
    An exemplary video is located in section ``Exercise 3: Bi-manual Coordination'' at \url{https://europeanacademy.org/training-tools/lastt/}.
}
with autonomous endoscope guidance within the allocated time.
The completion time is shown in Fig.~\ref{fig:autonomous_lastt_completion_time}. 
The average completion time for the $40$ trials was $172\,$s (only one outlier exceeding $300\,$s). 
As shown in the figure, the completion time for the \textit{plateau novices} was relatively constant. 
This was not the case for \textit{novices} and \textit{surgeons}, where a learning curve can be appreciated despite the initial 5--10 minutes of practice. 
The average completion time across participants decreased from $209\,$s in the first attempt to $144\,$s in the last exercise. 
These results indicate that the system provided repeatable behaviour that participants were able to learn.
}

 \section{Conclusion}
\label{sec:conclusion}

\rebmod{In this work we proposed the use of \emph{semantically rich instructions} to govern the interaction between a robotic autonomous endoscope holder and the operating surgeon. These are instructions such as ``focus on the right tool'' or ``focus the camera between the instruments''. This opposes previous endoscope holders handled via commands such as ``move up'' or ``zoom in''. 
\emph{Semantically rich instructions} are similar to the instructions surgeons would issue to a human camera operator, and can therefore be naturally adopted in clinical practice.
Thus, we believe that they may be a powerful tool to increase clinical acceptance.
}

\rebmod{As a first step towards implementing these instructions within a robotic endoscope holder, we concentrated our efforts on \emph{semantically rich instructions} related to surgical instruments, which we called \emph{autonomous instrument tracking (AIT)} instructions.
}
\rebnew{To implement these instructions we built a robotic system capable of executing them without the need for additional sensors besides the endoscope.
To the best of our knowledge, we are the first to report 
how to construct an autonomous instrument tracking system that allows for solo-surgery using only the endoscope as a sensor to track the instruments. 
Within the proposed system we included a novel tooltip detection method and a new 
visual servoing approach for a generalized endoscope model with support for remote center of motion and endoscope bending.
}

\rebnew{We found that our proposed localization method was able to detect tips in $84.46$\% of the frames, which in combination with our visual servoing approach allowed for a robust autonomous guidance of the endoscope.
With regards to the visual servoing method, we found that a hybrid of position-based visual servoing (PBVS) and 3D image-based visual-servoing (IBVS) is preferred for robotic endoscope control. 
}

\rebmod{During our experimental campaign we found that the REC-enabled AIT instructions yielded a predictable behaviour of the robotic endoscope holder that could be quickly understood and learned by the participants. The participants were able to execute a proven bi-manual coordination task within the prescribed completion time while assisted by the robotic endoscope holder. In three of the exercise runs, it was observed that the comanipulation fallback mode was required to solve for situations in which the instruments moved out of the view and the operator was unable to recover them in the view. This comanipulation mode thus ensures that failures in which the robotic endoscope holder has to be abandoned can be dealt with swiftly. An additional instruction to move back the robotic endoscope holder to a safe overview position could be considered as well. Such a safe location could for instance be close to the remote centre of motion (at the incision point). Although for  the general case, when flexible instruments are used, care should be paid that such retraction does not cause the bending segment to hinge behind anatomic structures.  
}

Besides the framework evaluation already performed, an in-depth comparison between human and robotic endoscope control remains as future work. Aspects such as time of completion, smoothness of motions, the stability of the image, number of corrections to the target zone, and average position of the instruments in the view remain to be compared. This contrast would quantify the difference in navigation quality between the proposed framework and a human-held endoscope.

\rebmod{While AIT instructions are necessary in most laparoscopic procedures, they are not the only instructions required for a semantic control of the endoscope holder, and it is a limitation of this study that it only focused on them.
Therefore, we are positive that this work will pave the way for further developments to enlarge the set of \emph{semantically rich instructions}.
} 
\section*{Acknowledgments}
This work was supported by core and project funding from the Wellcome/EPSRC [WT203148/Z/16/Z; NS/A000049/1; WT101957; NS/A000027/1]. This project has received funding from the European Union's Horizon 2020 research and innovation programme under grant agreement No 101016985 (FAROS project). T. Vercauteren is supported by a Medtronic/Royal Academy
of Engineering Research Chair [RCSRF1819\textbackslash7\textbackslash34].

\FloatBarrier

\appendix
\section*{Appendix}
\section{Inverse kinematics solution to PBVS}
\label{app:inverse_kinematics}

The inverse kinematics problem~\eqref{eq:inv_kin} can be solved analytically to obtain $(\theta^*_1,\theta^*_2,l^*)$. This problem has four possible solutions. To select the appropriate solution, it is important that the $z$-axis of $\{i\}$ is defined as the inward-pointing normal of the body wall.
As a first step, \eqref{eq:inv_kin} should be rewritten as:

\begin{equation}
    \vec{f}(\theta^*_1,\theta^*_2,l^*) = \left[\begin{matrix}f_x(\theta^*_1,\theta^*_2,l^*) \\ f_y(\theta^*_1,\theta^*_2,l^*) \\ f_z(\theta^*_1,\theta^*_2,l^*)\end{matrix}\right] = {\mat{T}^i_c}^* \tilde{\vec{s}}^* - \mat{T}^i_c \tilde{\vec{s}} = \vec{0}.
\end{equation}
Next, $l^*$ needs to be extracted from each expression in $(f_x, f_y, f_z)$, yielding respective expressions $(l_x^*, l_y^*, l_z^*)$. Equating 
$l^*_x = l^*_z$
and rewriting the result, eliminates $\theta^*_2$ and an expression of the form

\begin{equation}
    a_1 \sin(\theta^*_1) + b_1 \cos(\theta^*_1) = c_1
\end{equation}
emerges, with $a_1,b_1,c_1$ constants. Solving this for $\theta^*_1$ yields two supplementary angles, of which the solution with the smallest absolute value should be retained. If the expression $l_y^*$ is substituted in $f_x$ and $f_z$, and both are squared and added according to:

\begin{equation}
    f_x(\theta^*_1,\theta^*_2,l^*_y)^2 + f_z(\theta^*_1,\theta^*_2,l^*_y)^2 = 0,
\end{equation}
the dependence on $\theta^*_1$ cancels out. Simplifying this equation leads to:

\begin{equation}
    a_2 \cos^2(\theta^*_2) + b_2 \cos(\theta^*_2) + c_2 = 0,
\end{equation}
with $a_2,b_2,c_2$ constants. This is a quadratic equation in $\cos(\theta^*_2)$. The solution with the smallest $|\theta^*_2|$ is to be retained, but the sign of $\theta^*_2$ still needs to be confirmed. 
It is now possible to determine $l^*$, by plugging the known $\theta^*_1$ and $|\theta^*_2|$ into one of the expressions $(f_x,f_y,f_z)$. For numerical stability, $f_y$ should be used if $|\sin(\theta^*_2)| > \frac{1}{2}$, $f_x$ if $|\sin(\theta^*_1)| > \frac{1}{2}$, and $f_z$ otherwise. 
As the final step, the two unused expressions within $(f_x,f_y,f_z)$ need to be evaluated to determine the sign of $\theta^*_2$. If they do not evaluate to $0$, $\theta^*_2$ has to be negative and $l^*$ needs to be recomputed.

\bibliographystyle{model2-names.bst}\bibliography{library,more_refs}

\end{document}